\documentclass[10pt]{article} %
\usepackage[accepted]{tmlr}

\usepackage{amsmath,amsfonts,bm}

\def\eqref#1{equation~\ref{#1}}

\def\1{\bm{1}}

\DeclareMathAlphabet{\mathsfit}{\encodingdefault}{\sfdefault}{m}{sl}
\SetMathAlphabet{\mathsfit}{bold}{\encodingdefault}{\sfdefault}{bx}{n}

\usepackage{hyperref}
\usepackage{url}

\usepackage{graphicx}
\usepackage{subfigure}
\usepackage{booktabs}
\usepackage{enumitem}
\usepackage{amsmath}
\usepackage{xcolor}         %
\definecolor{Gray}{gray}{0.9}
\definecolor{White}{gray}{1.0}
\usepackage{colortbl} %
\usepackage{multirow}
\usepackage{wrapfig}
\usepackage{xkcdcolors}
\usepackage{algorithm}
\usepackage{algpseudocode}
\hypersetup{
    colorlinks=true,
    linkcolor=xkcdRed,
    urlcolor=xkcdBluish,
    linktoc=all,
    citecolor=xkcdOcean
}%

\newcommand\cifarwidth{32}
\newcommand\imagenetwidth{62}
\newcommand\ours{Score-Based Adversarial Generation}
\newcommand\oursacro{ScoreAG}
\newcommand{\D}{\mathrm{d}}
\newcommand{\change}[1]{{#1}}

\title{Assessing Robustness via Score-Based Adversarial Image Generation}

\author{\name Marcel Kollovieh{\normalfont\textsuperscript{1,2,3}}, Lukas Gosch{\normalfont\textsuperscript{1,2,3}}, Marten Lienen{\normalfont\textsuperscript{1}}, Yan Scholten{\normalfont\textsuperscript{1,2}}, Leo Schwinn{\normalfont\textsuperscript{1,2}}, Stephan Günnemann{\normalfont\textsuperscript{1,2,3}} \email m.kollovieh@tum.de \\
      \addr \textsuperscript{1}School of Computation, Information and Technology, Technical University of Munich, \textsuperscript{2}Munich Data Science Institute, \textsuperscript{3}Munich Center for Machine Learning
      }

\def\month{11}  %
\def\year{2024} %
\def\openreview{\url{https://openreview.net/forum?id=7Oqb6zlGWl}} %

\begin{document}

\maketitle

\begin{abstract}
  Most adversarial attacks and defenses focus on perturbations within small $\ell_p$-norm constraints. However, $\ell_p$ threat models cannot capture all relevant semantics-preserving perturbations, and hence, the scope of robustness evaluations is limited. In this work, we introduce Score-Based Adversarial Generation (ScoreAG), a novel framework that leverages the advancements in score-based generative models to generate unrestricted adversarial examples that overcome the limitations of $\ell_p$-norm constraints. Unlike traditional methods, ScoreAG maintains the core semantics of images while generating adversarial examples, either by transforming existing images or synthesizing new ones entirely from scratch. We further exploit the generative capability of ScoreAG to purify images, empirically enhancing the robustness of classifiers. Our extensive empirical evaluation demonstrates that ScoreAG improves upon the majority of state-of-the-art attacks and defenses across multiple benchmarks. This work highlights the importance of investigating adversarial examples bounded by semantics rather than $\ell_p$-norm constraints. ScoreAG represents an important step towards more encompassing robustness assessments.
\end{abstract}

\section{Introduction}
\begin{figure*}[h!]
    \centering
    \subfigure[Original]{\includegraphics[width=0.235\textwidth]{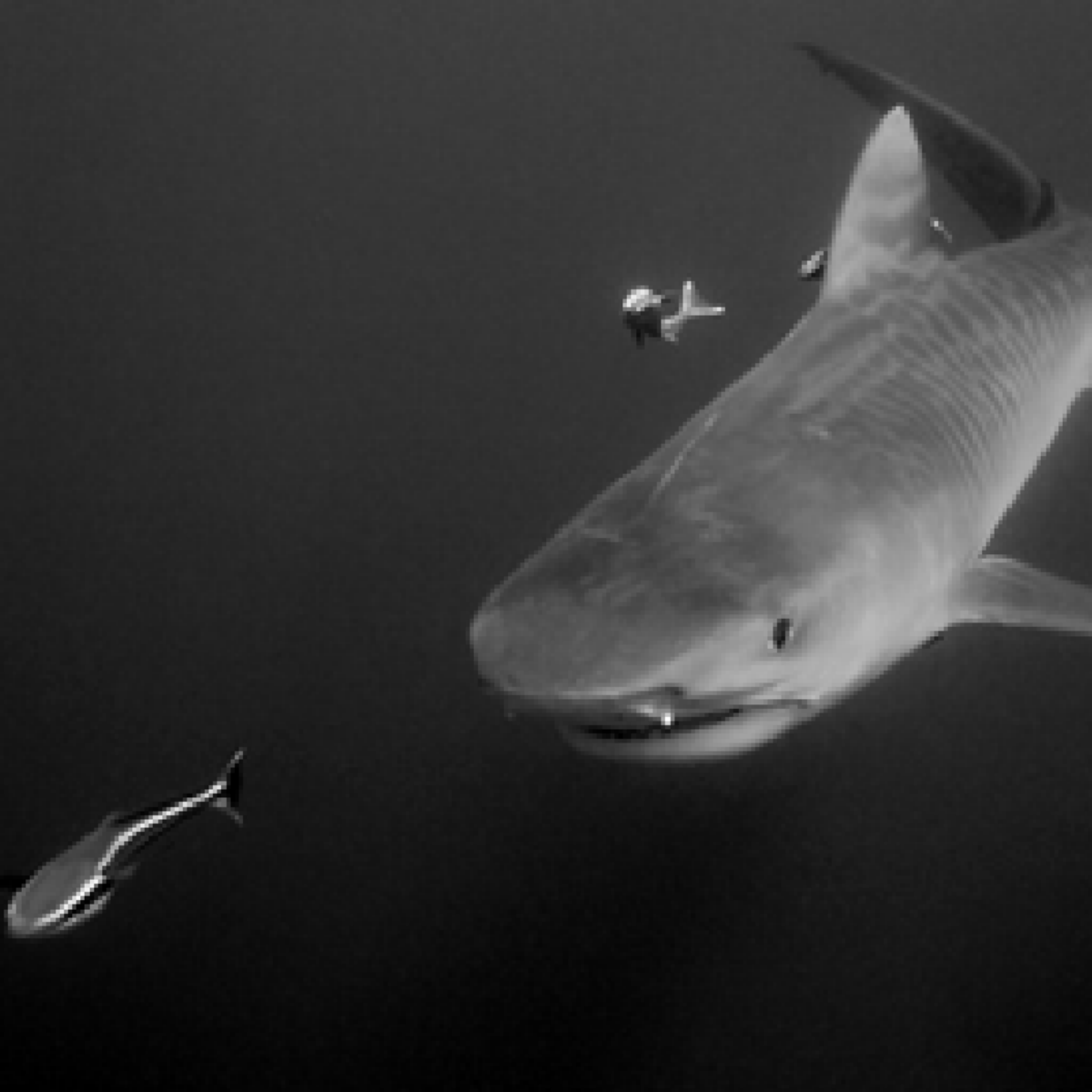}} 
    \subfigure[\oursacro{}-GAT (Ours)]{
\includegraphics[width=0.235\textwidth]{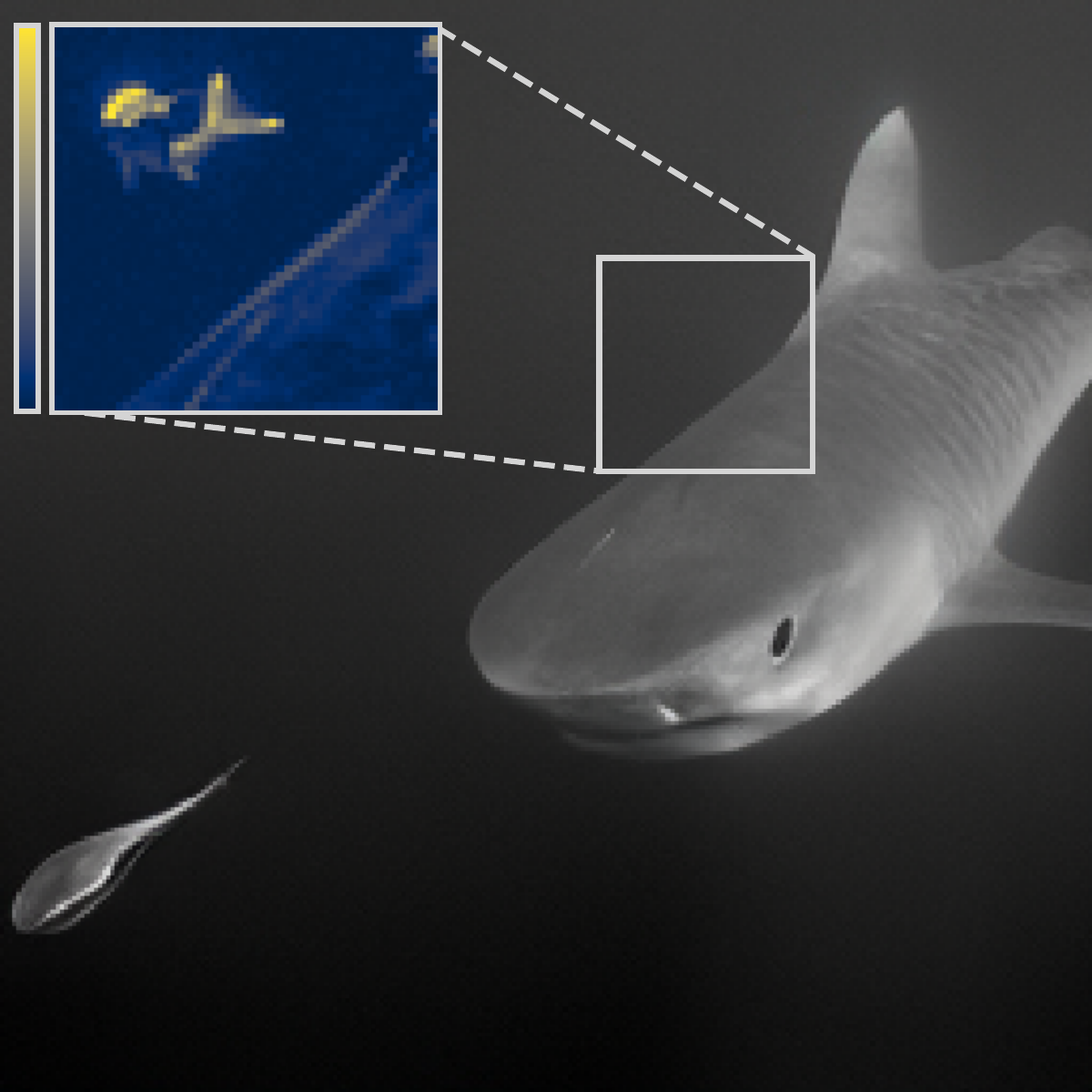}} 
    \subfigure[APGD ($\ell_2$)]{
\includegraphics[width=0.235\textwidth]{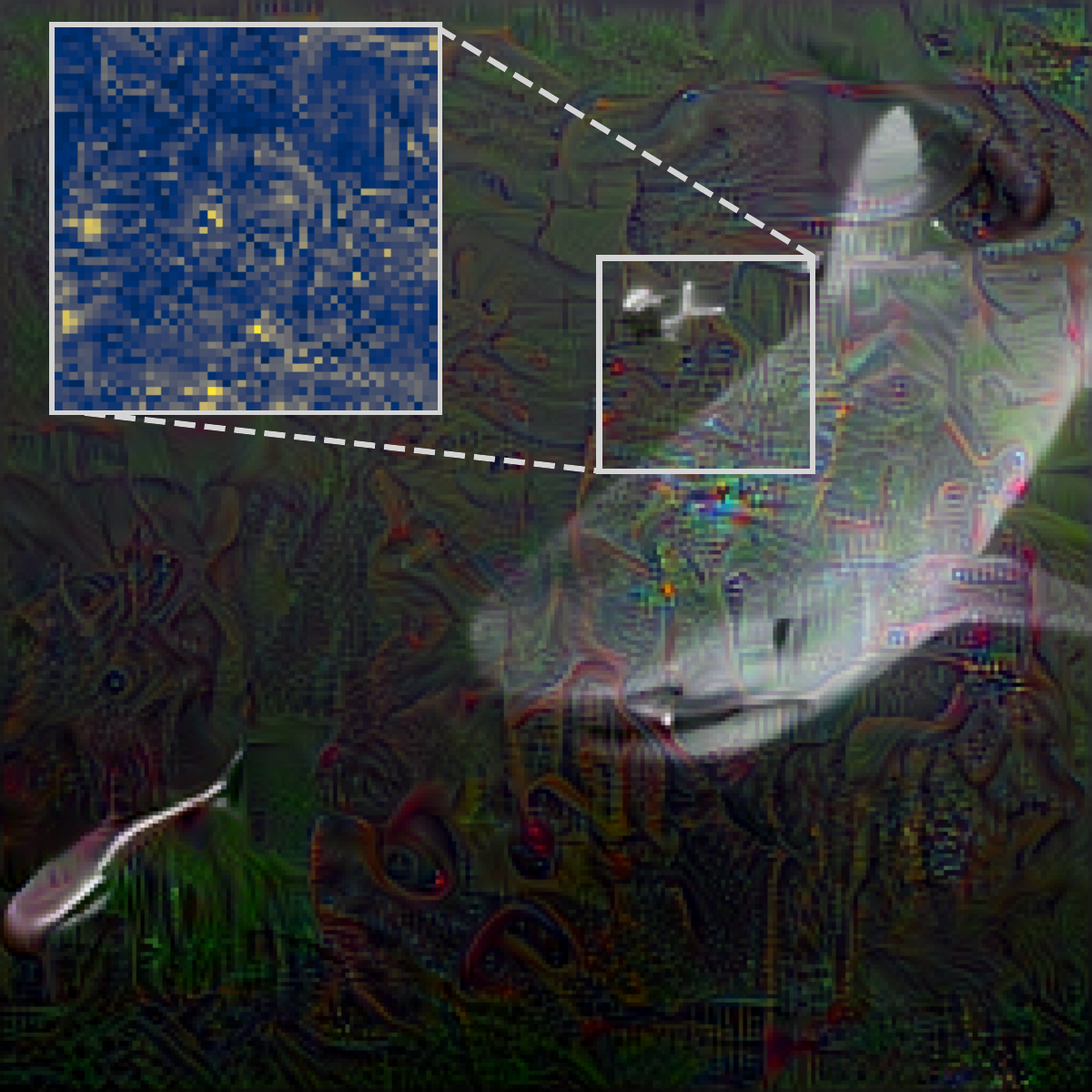}}
    \subfigure[APGD ($\ell_\infty$)]{
\includegraphics[width=0.235\textwidth]{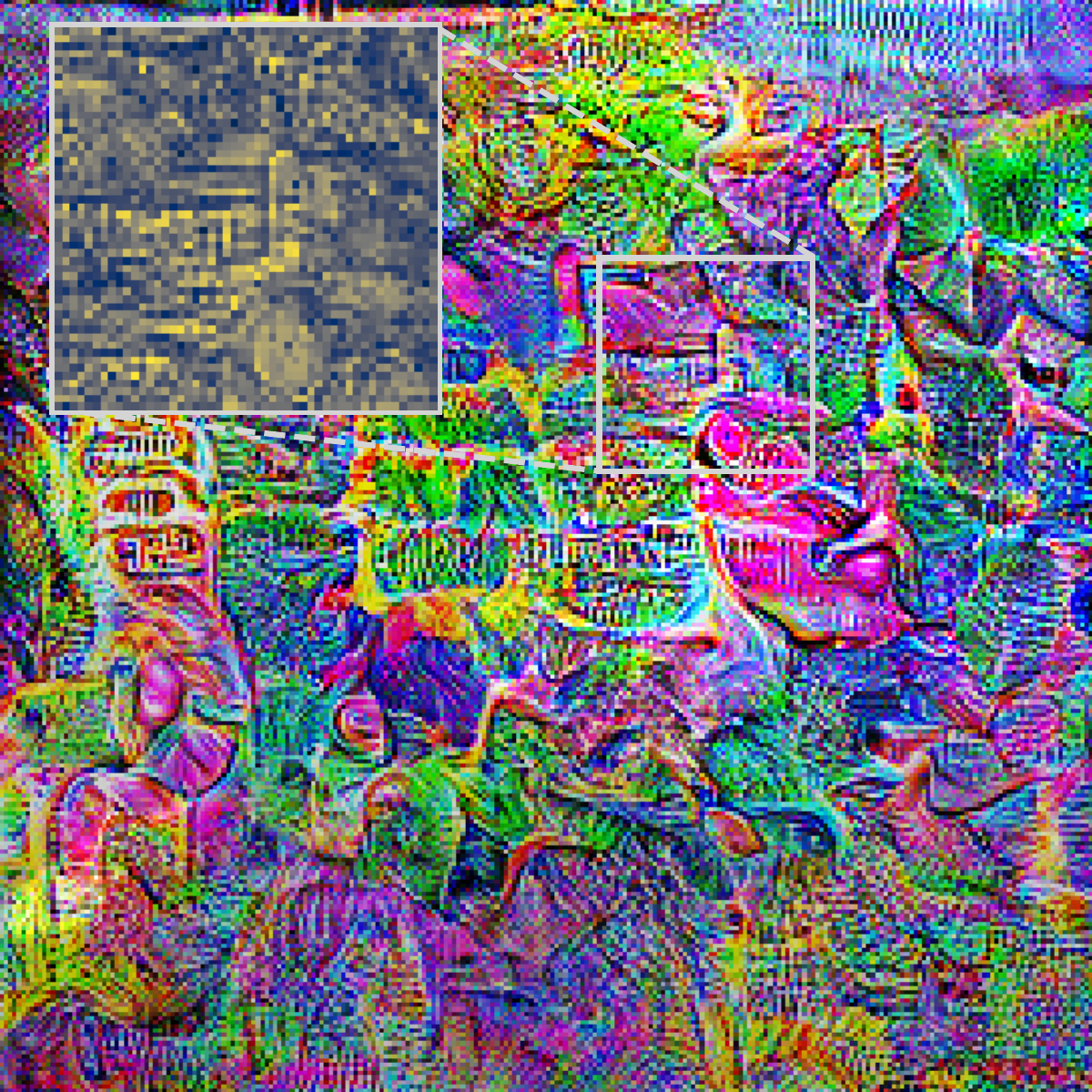}}
    \vspace{-0.5em}
    \caption{Examples of various adversarial attacks on an image of the class ``tiger shark" (a). The inset visualizes a heatmap of the strength of the corresponding perturbation. Despite the fact that the perturbation generated by ScoreAG-GAT (b) lies outside of common $\ell_p$-norm constraints ($\ell_\infty=188/255$, $\ell_2=18.47$), it is aware of the semantics: removing a small fish to change the predicted label to ``hammer shark". This is in stark contrast to APGD \citep{croce2020reliable} with matching norm constraints, which either (c) results in highly perceptible and unnatural changes, or (d) fails to preserve image semantics completely. \change{This is an example of Generative Adversarial Transformation (GAS), one of the three use-cases of \oursacro{}.}}
    \label{fig:main}
\end{figure*}

Ensuring robustness %
against noisy data or malicious interventions has become a major concern in various applications ranging from autonomous driving~\citep{eykholt2018physical} and medical diagnostics~\citep{dong2023adversarial} to the financial sector~\citep{fursov2021adversarial}. Even though adversarial robustness has received significant research attention \citep{goodfellow2014explaining, madry2017towards, croce2020reliable}, it is still an unsolved problem. %
Most works on adversarial robustness define adversarial perturbations to lie within a restricted $\ell_p$-norm from the input.
However, recent works have shown that significant semantic changes can occur within common perturbation norms, and that many relevant semantics-preserving corruptions lie outside specific norm ball choices ~\citep{tramer2020fundamental, gosch2023revisiting}. 
Examples include physical perturbations such as stickers on stop signs \citep{eykholt2018physical} or natural %
corruptions such as lighting or fog \citep{kar20223d, hendrycks2018benchmarking}.
This led to the inclusion of a first $\ell_p$-norm independent robustness benchmark to RobustBench \citep{croce2020robustbench}, and a call to further investigation into robustness beyond $\ell_p$-bounded adversaries \citep{hendrycks2022unsolved}. Thus, in this work, we address the following research question: 
\vspace{-0.75em}
\begin{center}
\textit{How can we generate semantics-preserving adversarial examples beyond 
$\ell_p$-norm constraints?}    
\end{center}
\vspace{-0.75em}
We propose to leverage the significant progress in diffusion models \citep{sohl2015deep, ho2020denoising} and score-based generative models \citep{song2020score} in generating realistic images. Specifically, we introduce \emph{\ours{} (\oursacro{})}, a framework designed to synthesize adversarial examples, transform existing images into adversarial ones, and purify images.  
Using diffusion guidance~\citep{dhariwal2021diffusion}, \oursacro{} can generate semantics-preserving adversarial examples that are not captured by common $\ell_p$-norms (see Fig.~\ref{fig:main}).
Overall, \oursacro{} represents a novel tool for assessing and enhancing the empirical robustness of image classifiers.

Our \textit{key contributions} are summarized as follows:
\vspace{-0.3cm}
\begin{itemize}[itemsep=0.5pt, leftmargin=*]
    \item We overcome limitations of $\ell_p$ threat models by proposing \oursacro{}, a framework utilizing diffusion guidance on pre-trained models, enabling the generation of \emph{unrestricted} but semantics-preserving adversarial examples. %
    \item With \oursacro{} we \emph{transform} existing 
    images into adversarial ones as well as \emph{synthesize} completely new adversarial examples.
    \item We show that \oursacro{} enhances classifier robustness by \emph{purifying} adversarial examples and common corruptions.
    \item We demonstrate \oursacro{}'s capability
    in an exhaustive empirical evaluation and show it is able to outperform a majority of existing attacks and defenses on several benchmarks. Additionally, we underscore \oursacro{}'s semantic preserving ability in a human study.
\end{itemize}

\section{Background}\label{sec:background}
\textbf{Score-Based Generative Modelling.}
Score-based generative models \citep{song2020score} are a class of generative models based on a continuous-time diffusion process $\{\mathbf{x}_t\}_{t\in[0,1]}$ accompanied by their corresponding probability densities $p_t(\mathbf{x})$. The diffusion process progressively perturbs a data distribution $\mathbf{x}_0~\sim p_0$ into a prior distribution $\mathbf{x}_1\sim p_1$. This transformation is formalized as a Stochastic Differential Equation (SDE), i.e.,
\begin{align}
    \D\mathbf{x}_t=\mathbf{f}(\mathbf{x}_t, t)\D t+ g(t)\D \mathbf{w}, \label{eq:sde}
\end{align}
where $\mathbf{f}(\cdot, t):\mathbb{R}^d\rightarrow\mathbb{R}^d$ represents the drift coefficient of $\mathbf{x}_t$, $g(\cdot):\mathbb{R}\rightarrow\mathbb{R}$ the diffusion coefficient, and $\mathbf{w}$ the standard Wiener process (i.e., Brownian motion). Furthermore, let $p_{st}(\mathbf{x}_t\mid\mathbf{x}_s)$ describe the transition kernel from $\mathbf{x}_s$ to $\mathbf{x}_t$, where $s<t$. 

By appropriately choosing $\mathbf{f}$ and $g$, $p_1$ asymptotically converges to an isotropic Gaussian distribution, i.e., $p_1\approx\mathcal{N}(\mathbf{0},\mathbf{I})$. To generate data, the reverse-time SDE needs to be solved:
\begin{align}
    \D\mathbf{x}_t=[\mathbf{f}(\mathbf{x}_t, t) - g(t)^2\nabla_{\mathbf{x}_t}\log p_t(\mathbf{x}_t)] \D t + g(t)\D\mathbf{w}\label{eq:reverse}.
\end{align}
Solving the SDE requires access to the time-dependent score function $\nabla_{\mathbf{x}_t}\log p_t(\mathbf{x}_t)$, which is typically unknown. Instead, the score function is estimated using a neural network $\mathbf{s}_\theta(\mathbf{x}_t,t)$. The parameters of this network are learned by minimizing the following cost function:
\begin{equation}
\begin{aligned}
\mathbb{E}_t \Big [\lambda(
t) \mathbb{E}_{\mathbf{x}_0}\mathbb{E}_{\mathbf{x}_t\mid \mathbf{x}_0} \left[ \|\mathbf{s}_{\theta}(\mathbf{x}_t, t) - \nabla_{\mathbf{x}_t} \log p_{0t}(\mathbf{x}_t\mid \mathbf{x}_0)\|^2_2 \right] \Big]. \label{eq:loss}
\end{aligned}
\end{equation}
Here, $\lambda(\cdot):[0,1]\rightarrow\mathbb{R}_{>0}$ serves as a time-dependent weighting parameter, and $t$ is uniformly sampled from the interval $[0,1]$.

In this formulation, $\mathbf{x}_0 \sim p_0$ is sampled from the data distribution, and $\mathbf{x}_t \sim p_{0t}(\mathbf{x}_t\mid\mathbf{x}_0)$ follows the diffusion process at time $t$. The goal is to train the network $\mathbf{s}_\theta$ to accurately match the true score function $\nabla_{\mathbf{x}_t} \log p_{0t}(\mathbf{x}_t\mid\mathbf{x}_0)$, enabling data generation through the reverse diffusion process, which can be solved using numerical solvers.

\textbf{Diffusion Guidance.}
To enable conditional generation with unconditionally trained diffusion models, \citet{dhariwal2021diffusion} introduce classifier guidance. The central idea is to generate samples from the conditional distribution $p(\mathbf{x}_0\mid c)$, where $c$ represents a specific class, i.e., sampling images of class $c$. To achieve this, the authors replace the gradient of the unconditional distribution $p_t(\mathbf{x}_t)$ in the reverse process (see \eqref{eq:reverse}) with its conditional counterpart. 

By applying Bayes' theorem, the gradient of the conditional gradient can be decomposed as:
\begin{align}
    \nabla_{\mathbf{x}_t} \log p(\mathbf{x}_t\mid c) = \nabla_{\mathbf{x}_t} \log p(\mathbf{x}_t) + \nabla_{\mathbf{x}_t} \log p(c\mid \mathbf{x}_t),
\end{align}
where $\nabla_{\mathbf{x}_t} \log p(\mathbf{x}_t)$ represents unconditional score function and $\nabla_{\mathbf{x}_t} \log p(c\mid \mathbf{x}_t)$ represents the guidance score. The unconditional score function is approximated using the neural network $\mathbf{s}_{\theta}$, which is trained using the loss in \eqref{eq:loss}. 

To compute the guidance score $\nabla_{\mathbf{x}_t} \log p(c\mid \mathbf{x}_t)$, \citet{dhariwal2021diffusion} utilize the gradients of a time-dependent classifier $f(\mathbf{x}_t, t)$ with respect to $\mathbf{x}_t$. The guidance score steers the generation process towards samples that are consistent with the desired class $c$. This method allows an unconditional diffusion model, i.e., a model trained without conditional information, to be adapted for conditional tasks, enabling the generation of class-specific samples.

Classifier guidance has since been extended to handle arbitrary conditions $c$, such as guiding generation towards CLIP embeddings \citep{nichol2021glide}. This flexibility in choosing different conditions is essential to \oursacro{} and enables us to adapt the model for three distinct tasks by adjusting the guidance condition, as described in the next section.

\section{\ours{}}\label{sec:methodology}
\begin{figure*}
    \centering
    \includegraphics[width=0.94\textwidth]{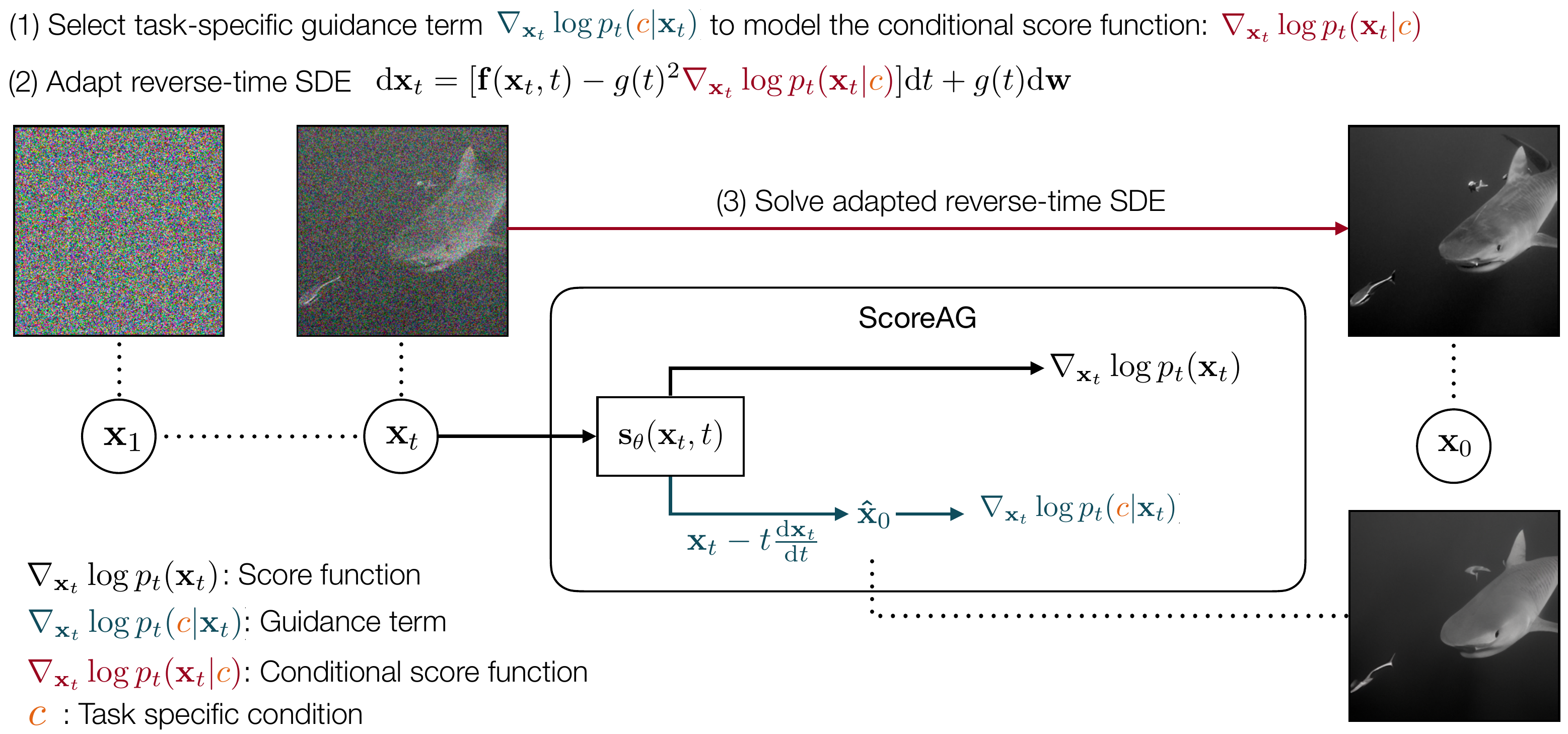} %
    \vspace{-0.75em}
    \caption{An overview of \oursacro{} and its three steps. \oursacro{} starts from noise $\mathbf{x}_1$ and iteratively denoises it into an image $\mathbf{x}_0$. It uses the task-specific guidance terms $\nabla_{\mathbf{x}_t} \log p_t(c\mid\mathbf{x}_t)$ and the score function $\nabla_{\mathbf{x}_t}\log p_t(\mathbf{x}_t)$ to guide the process towards the task specific condition $c$. The network $\mathbf{s}_\theta$ is used for approximating the score function $\nabla_{\mathbf{x}_t}\log p_t(\mathbf{x}_t)$ and for the one-step Euler prediction $\hat{\mathbf{x}}_0$.}
    \vspace{-1.25em}
    \label{fig:methodology}
\end{figure*}
In this section, we introduce \emph{\ours{}} (\oursacro{}), a framework employing generative models to evaluate robustness beyond the $\ell_p$-norm constraints. \oursacro{} is designed to perform the following three tasks: \textbf{(1)} the generation of adversarial images (see Sec.~\ref{sec:synthesize}), \textbf{(2)} the transformation of existing images into adversarial examples (see Sec.~\ref{sec:transform}), and \textbf{(3)} the purification of images to enhance empirical robustness of classifiers (see Sec.~\ref{sec:purification}).

\oursacro{} consists of three steps: \textbf{(1)} select a guidance term for the corresponding task to model the conditional score function $\nabla_{\mathbf{x}_t} \log p(\mathbf{x}_t\mid c)$, (\textbf{2}) adapt the reverse-time SDE with the task-specific conditional score function, and \textbf{(3)} solve the adapted reverse-time SDE for an initial noisy image $\mathbf{x}_1\sim\mathcal{N}(\mathbf{0},\mathbf{I})$ using numerical methods. Depending on the task, the result is either an adversarial or a purified image. We provide an overview of \oursacro{} in Fig.~\ref{fig:methodology}.

In detail, the conditional score function is composed of the normal score function $\nabla_{\mathbf{x}_t} \log p_t(\mathbf{x}_t)$ and the task-specific guidance term $\nabla_{\mathbf{x}_t} \log p(c\mid\mathbf{x}_t)$, that is 
\begin{align}
    \nabla_{\mathbf{x}_t}\log p_t(\mathbf{x}_t\mid c)=\nabla_{\mathbf{x}_t} \log p_t(\mathbf{x}_t) + \nabla_{\mathbf{x}_t} \log p_t(c\mid\mathbf{x}_t)\label{eq:guidance},
\end{align}
where $\log p_t(\mathbf{x}_t)$ is modeled by a score-based generative model. Solving the adapted reverse-time SDE yields a sample of the conditional distribution $p(\mathbf{x}_0\mid c)$, i.e., an adversarial or purified image. To simplify the presentation, we will denote class-conditional functions as \smash{$p_y(\mathbf{x}_t)$} rather than the more verbose \smash{$p(\mathbf{x}_t\mid y)$}.
\subsection{Problem Statement.}
In the realm of adversarial robustness, traditional evaluation methods often constrain adversarial perturbations within an $\ell_p$-norm ball, providing a limited robustness assessment. These limitations are addressed by \textit{unrestricted} attacks. In this work, we consider the following three key tasks: \textbf{(1)} Generating new adversarial images that inherently belong to a specific class $y^*$ but are misclassified by the classifier as $\tilde{y}$; \textbf{(2)} Transforming existing images $\mathbf{x}^*$ into adversarial examples, i.e., images that are misclassified as $\tilde{y}$ (see adversary) while maintaining their core semantics and true class $y^*$; and \textbf{(3)} Purifying adversarial images $\mathbf{x}_\mathrm{ADV}$ to recover correct classification and enhance empirical robustness.

\textbf{Adversary.} Let $y^{*}\in\{1,\dots, K\}$ denote the true class of a clean image $\mathbf{x}\in[0,1]^{C\times H\times W}$, $\tilde{y}\neq y^*$ be a different class, and $f(\cdot):[0,1]^{C\times H\times W}\rightarrow \{1,\dots,K\}$ a classifier. An image $\mathbf{x}_\mathrm{ADV}\in[0,1]^{C\times H\times W}$ is termed an adversarial example if it is misclassified by $f$, i.e., $f(\mathbf{x})=y^*\neq \tilde{y}=f(\mathbf{x}_{\mathrm{ADV}})$,
while preserving the semantics, i.e.,  $\Omega(\mathbf{x})=\Omega(\mathbf{x}_\mathrm{ADV})$ with $\Omega$ denoting a semantics-describing oracle. Therefore, adversarial examples do not change the true label of the image. To enforce this, conventional adversarial attacks restrict the perturbation to lie in a certain $\ell_p$-norm, avoiding large differences to the original image. In contrast, \oursacro{} is not limited by $\ell_p$-norm restrictions but preserves the semantics by employing a class-conditional generative model. In the following, we introduce each task in detail.

\subsection{Generative Adversarial Synthesis}\label{sec:synthesize}
Generative Adversarial Synthesis (GAS) aims to synthesize images that are adversarial by nature. While these images maintain the semantics of a certain class $y^*$, they are misclassified by a classifier into a different class $\tilde{y}$. The formal objective of GAS is to sample from the distribution $p_{y^*}(\mathbf{x}_0\mid f(\mathbf{x}_0)=\tilde{y})$, where $f(\mathbf{x}_0)=\tilde{y}$ corresponds to the guidance condition $c$.

Applying Bayes' theorem according to \eqref{eq:guidance}, the conditional score can be expressed as:
\begin{equation}
    \begin{aligned}
        \nabla_{\mathbf{x}_t}\log p_{t,y^*}(\mathbf{x}_t\mid f(\mathbf{x}_0)= \tilde{y}) = \nabla_{\mathbf{x}_t}\log p_{t,y^*}(\mathbf{x}_t)+ s_\mathbf{y}\cdot\nabla_{\mathbf{x}_t}\log p_{t,y^*}(f(\mathbf{x}_0)=\tilde{y}\mid\mathbf{x}_t),
    \end{aligned}
    \label{eq:synthesis}
\end{equation}
where $s_\mathbf{y}$ is a scaling parameter adjusting the strength of the attack.
While \smash{$\nabla_\mathbf{x}\log p_{t,y^*}({\mathbf{x}_t})$} can be learned with a class-conditional score network \smash{$\mathbf{s}_\theta(\mathbf{x}_t,t,y)$}, \smash{$\nabla_{\mathbf{x}_t}\log p_{t,y^*}(f(\mathbf{x}_0) = \tilde{y} \mid \mathbf{x}_t)$} requires further analysis. By marginalizing over $\mathbf{x}_0$ and using the Markov property that $f(\mathbf{x}_0)$ and $\mathbf{x}_t$ are independent given $\mathbf{x}_0$, we see that
\begin{align}
    p_{t,y^*}(f(\mathbf{x}_0)=\tilde{y}\mid\mathbf{x}_t) = \mathbb{E}_{\mathbf{x}_0\mid\mathbf{x}_t, t, y^*}[p(f(\mathbf{x}_0)=\tilde{y}\mid\mathbf{x}_0)] \label{eq:expected}
\end{align}
is the expected probability of classifying generated samples $\mathbf{x}_0$ as class $\tilde{y}$.

While a direct Monte Carlo approximation to \eqref{eq:expected} is theoretically feasible, drawing samples from the class-conditional generative model $p_{t,y^*}(\mathbf{x}_0 \mid \mathbf{x}_t)$ would be expensive.
Instead, we approximate $p_{t,y^*}(\mathbf{x}_0 \mid \mathbf{x}_t)$ as a Dirac distribution centered on the one-step Euler solution $\hat{\mathbf{x}}_0$ to \eqref{eq:sde} from $t$ to $0$
$
    \mathbf{\hat{x}}_0=\mathbf{x}_t-t\frac{\D \mathbf{x}_t}{\D t },
$
which simplifies \eqref{eq:expected} to
\begin{align}
    p_{t,y^*}(f(\mathbf{x}_0)=\tilde{y}\mid\mathbf{x}_t) \approx p(f(\hat{\mathbf{x}}_0) = \tilde{y} \mid \mathbf{x}_t). \label{eq:euler}
\end{align}

Thus, we approximate $\nabla_{\mathbf{x}_t}\log p_{t,y^*}(f(\mathbf{x}_0)=\tilde{y}\mid\mathbf{x}_t) \approx \nabla_{\mathbf{x}_t}\log p(f(\mathbf{\hat{x}}_0)=\tilde{y}\mid\mathbf{x}_t)$,
which, in practice, corresponds to maximizing the cross-entropy between the classification $f(\mathbf{\hat{x}}_0)$ of the generated sample and the target class $\tilde{y}$.

In contrast to \citet{dhariwal2021diffusion}, our approximation allows us to work with the classifier $f$ directly instead of fine-tuning a time-dependent variant. Moreover, this can be adapted to discrete-time diffusion models with the approach by \citet{kollovieh2023predict}. %

\subsection{Generative Adversarial Transformation}\label{sec:transform}
While in GAS we synthesize adversarial samples from scratch, Generative Adversarial Transformation (GAT) focuses on transforming existing images into adversarial examples. For a given image $\mathbf{x}^*$ and its corresponding true class label $y^*$, the objective is to sample a perturbed image misclassified as $\tilde{y}$ while preserving the core semantics of $\mathbf{x}^*$. We denote the resulting distribution as $p_{y^*}(\mathbf{x}_0\mid f(\mathbf{x}_0)=\tilde{y},\mathbf{x}^*)$ for the guidance condition $c=\{\mathbf{x}^*,f(\mathbf{x}_0)=\tilde{y}\}$ leading to the following conditional score (\eqref{eq:guidance}):
\begin{equation}
\begin{aligned}
  \nabla_{\mathbf{x}_t}\log p_{t,y^*}(\mathbf{x}_t\mid\mathbf{x}^*,f(\mathbf{x}_0)=\tilde{y}) = \nabla_{\mathbf{x}_t}\log p_{t,y^*}(\mathbf{x}_t) + \nabla_{\mathbf{x}_t}\log p_{t,y^*}(\mathbf{x}^*,&f(\mathbf{x}_0)=\tilde{y}\mid\mathbf{x}_t)\label{eq:transform}.
\end{aligned}
\end{equation}
By assuming independence between $\mathbf{x}^*$ and $\tilde{y}$ given $\mathbf{x}_t$, we split the guidance term into $s_\mathbf{x}\cdot\nabla_{\mathbf{x}_t}\log p_{t,y^*}(\mathbf{x}^*\mid\mathbf{x}_t)+s_\mathbf{y}\cdot\nabla_{\mathbf{x}_t}\log p_{t,y^*}(f(\mathbf{x}_0)=\tilde{y}\mid\mathbf{x}_t)$, implying that $\tilde{y}$ should not influence the core semantics of the given image. Note that we introduced the two scaling parameters $s_\mathbf{x}$ and $s_\mathbf{y}$ to control the possible deviation from the original image and the strength of the attack, respectively. While we treat the score function $\nabla_{\mathbf{x}_t}\log p_{t,y^*}(\mathbf{x}_t)$ and the guidance term $\nabla_{\mathbf{x}_t}\log p_{t,y^*}(f(\mathbf{x}_0)=\tilde{y}\mid\mathbf{x}_t)$ as in the GAS setup, we model the distribution $p_{t,y^*}(\mathbf{x}^*\mid\mathbf{x}_t)$ as a Gaussian centered at the one-step Euler prediction $\mathbf{\hat{x}}_0$ (\eqref{eq:euler}),
\begin{align}
    p_{t,y^*}(\mathbf{x}^*\mid\mathbf{x}_t)=\mathcal{N}(\mathbf{\hat{x}}_0,\mathbf{I}).\label{eq:obs}
\end{align}
It follows that our sampling process searches for an adversarial example while minimizing the squared error between $\mathbf{x}^*$ and $\hat{\mathbf{x}}_0$. Importantly, this lets us generate samples $\mathbf{x}_0$ close to $\mathbf{x}^*$ without imposing specific $\ell_p$-norm constraints. Furthermore, our framework is not limited to the squared error, but can also utilize other differentiable similarity metrics as guidance such as the LPIPS~\citep{zhang2018unreasonable} score. Note that, while adversarial examples generated by \oursacro{} are unrestricted in the sense of $\ell_p$-balls, they are constrained to the data manifold of the generative model through the construction of our generative process.
This yields an \emph{unrestricted} attack that preserves the core semantics using the class-conditional score network $\mathbf{s}_\theta$.

As a result, GAT provides a more comprehensive robustness assessment than traditional $\ell_p$-threat models. This enhanced assessment capability stems from the inherent properties of GAT, which (1) encompasses all semantics-preserving adversarial examples within the $\ell_p$-balls as captured by the generative model, and (2) includes semantics-preserving adversarial examples that conventional $\ell_p$-threat models do not capture.

\subsection{Generative Adversarial Purification}\label{sec:purification}
Generative Adversarial Purification (GAP) extends \oursacro{} to counter adversarial attacks. It is designed to purify adversarial images, i.e., remove adversarial perturbations through its generative capability to enhance the robustness of machine learning models. 

Given an adversarial image $\mathbf{x}_{\mathrm{ADV}}$ that was perturbed to induce a misclassification, GAP aims to sample an image from the data distribution that resembles the semantics of $\mathbf{x}_\mathrm{ADV}$, which we denote as $p(\mathbf{x}_0\mid\mathbf{x}_\mathrm{ADV})$ with $\mathbf{x}_{\mathrm{ADV}}$ corresponding to the guidance condition $c$. We model its score function analogously to \eqref{eq:transform}:
\begin{equation}
\begin{aligned}
  \nabla_{\mathbf{x}_t}\log p_t(\mathbf{x}_t\mid\mathbf{x}_{\mathrm{ADV}}) = \nabla_{\mathbf{x}_t}\log p_t(\mathbf{x}_t)+s_\mathbf{x}&\cdot\nabla_{\mathbf{x}_t}\log p_t(\mathbf{x}_{\mathrm{ADV}}\mid\mathbf{x}_t),
\end{aligned}
\label{eq:purification}
\end{equation}
where $s_\mathbf{x}$ is a scaling parameter controlling the deviation from the input.
Note that we omit $y^*$ since there is no known ground-truth class label. 
As previously, we utilize a time-dependent score network $\mathbf{s}_\theta$ to approximate the term $\nabla_{\mathbf{x}_t}\log p_t(\mathbf{x}_t)$. The term $p_t(\mathbf{x}_{\mathrm{ADV}}\mid\mathbf{x}_t)$ is modeled according to \eqref{eq:obs}, as before assuming it follows a Gaussian distribution with a mean of the one-step Euler prediction $\mathbf{\hat{x}}_0$. Note that \oursacro{}, just as other purification methods, cannot detect adversarial images. Therefore, \change{it also needs} to preserve image semantics if there is no perturbation. 

\section{Experimental Evaluation}\label{sec:experiments}
The primary objective of our experimental evaluation is to assess the capability of \oursacro{} in generating \change{and purifying} adversarial examples. More specifically, we investigate the following properties of \oursacro{}: \textbf{(1)} the ability to synthesize adversarial examples from scratch (GAS), \textbf{(2)} the ability to transform existing images into adversarial examples (GAT), and \textbf{(3)} the enhancement of classifier robustness by leveraging the generative capability of the model to purify images (GAP). This evaluation aims to provide comprehensive insights into the strengths and limitations of \oursacro{} in the realm of adversarial example generation and classifier robustness.

\textbf{Baselines.}
In our evaluation, we benchmark our adversarial attacks against a wide range of established methods covering various threat models. Specifically, we consider the fast gradient sign-based approaches FGSM~\citep{goodfellow2014explaining}, DI-FGSM~\citep{xie2019improving}, and SI-NI-FGSM~\citep{lin2019nesterov}. In addition, we include %
Projected Gradient Descent-based techniques, specifically Adaptive Projected Gradient Descent (APGD) and its targeted variant (APGDT)~\citep{croce2020reliable}. For a comprehensive assessment, we also examine single pixel, black-box, and minimal perturbation methods, represented by OnePixel~\citep{su2019one}, Square~\citep{andriushchenko2020square} and Fast Adaptive Boundary (FAB)~\citep{croce2020minimally}, respectively. Finally, we compare to the unrestricted attacks 
Composite Adversarial Attack (CAA)~\citep{hsiung2022caa}, PerceptualPGDAttack (PPGD), FastLagrangePerceptualAttack (LPA)~\citep{laidlaw2020perceptual}, and DiffAttack~\citep{chen2023diffusion}, which is based on latent diffusion. Furthermore, we compare to Adversarial Content Attack (ACA)~\citep{chen2023content} in App.~\ref{sec:additional_classifiers}.\footnote{We do not compare to \citet{chen2023advdiffuser} as their code was not publicly available upon submission and multiple attempts to contact the authors were unsuccessful.} 

To evaluate the efficacy of \oursacro{} in purifying adversarial examples, we conduct several experiments in a preprocessor-blackbox setting. For the evaluation, we employ the targeted APGDT and untargeted APGD attacks~\citep{croce2020reliable} and \oursacro{} in the GAS setup. Our experiments also incorporate the purifying methods ADP~\citep{yoon2021adversarial} and DiffPure~\citep{nie2022DiffPure}. Additionally, we compare with state-of-the-art adversarial training techniques that partially utilize supplementary data from generative models \citep{cui2023decoupled, wang2023better, peng2023robust}.

\textbf{Experimental Setup.} We employ three benchmark datasets for our experiments: CIFAR10, CIFAR100 \citep{krizhevsky2009learning}, and TinyImagenet. We utilize pre-trained Elucidating Diffusion Models (EDM) in the variance preserving (VP) setup ~\citep{karras2022elucidating, wang2023better} for image generation. As our classifier, we opt for the well-established WideResNet architecture WRN-28-10~\citep{zagoruyko2016wide}. The classifiers are trained for 400 epochs using SGD with Nesterov momentum of $0.9$ and weight decay of $5\times 10^{-4}$. Additionally, we incorporate a cyclic learning rate scheduler with cosine annealing~\citep{smith2019super} with an initial learning rate of $0.2$. To further stabilize the training process, we apply exponential moving average with a decay rate of $0.995$. Each classifier is trained four times to ensure the reproducibility of our results\change{, and we report standard deviations with ($\pm$). For pretrained classifiers with only one available model, we do not report standard deviations.} For the restricted methods, we consider the common norms in the literature $\ell_2=0.5$ for CIFAR10 and CIFAR100, $\ell_2=2.5$ for TinyImagenet, and $\ell_\infty=8/255$ for all three datasets. For DiffAttack, ACA, and DiffPure we take the implementation of the official repositories, while we use Torchattacks~\citep{kim2020torchattacks} for the remaining baselines. \change{The runtimes for all methods are shown in Tab.~\ref{tab:runtimes}} in the appendix.%

\textbf{Evaluation Metrics.} %
To evaluate our results, we compute the robust accuracy, i.e., the accuracy after an attack. Furthermore, we use the clean accuracy, i.e., the accuracy of a (robust) model without any attack. For the GAS task, we use the FID \citep{heusel2017gans} to assess the similarity between the distribution of synthetic images and the test set, providing a distribution-level measure. Since FID is not suitable for instance-based evaluation, we use the LPIPS score \citep{zhang2018unreasonable} for the GAT task to measure perceptual similarity at the instance level.

\begin{figure*}[t]
    \centering
    \includegraphics[width=0.9\textwidth]{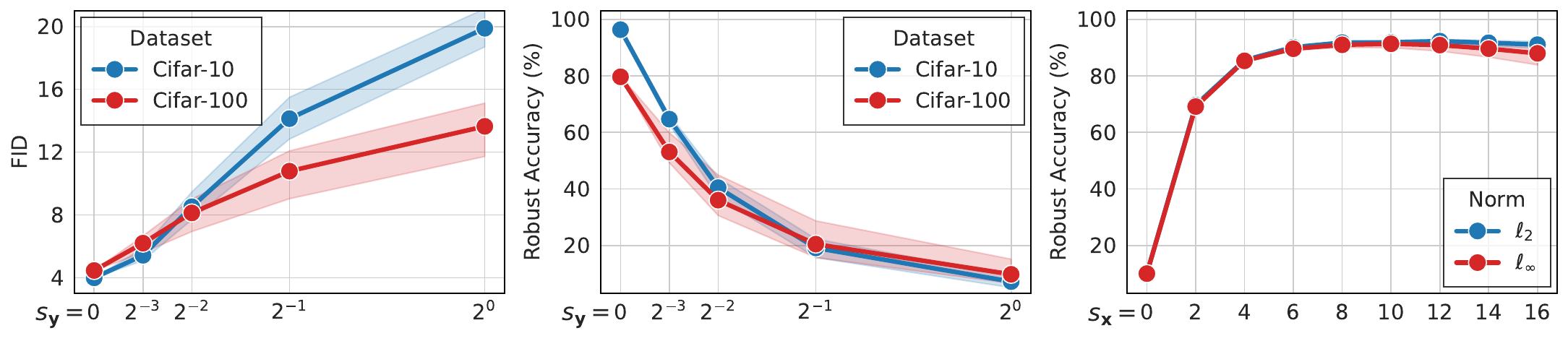}\vspace{-1.5em}
    \subfigure[\label{fig:scale_left} FID (GAS).]{\hspace{0.32\textwidth}}
    \subfigure[\label{fig:scale_middle} Robust Accuracy (GAS).]{\hspace{0.28\textwidth}}
    \subfigure[\label{fig:scale_right} Robust Accuracy (GAP).]{\hspace{0.28\textwidth}}
    \vspace{-0.75em}
    \caption{%
    FID (a) and accuracy (b) for increasing $s_\mathbf{y}$ scales in the synthesis (GAS) setup, and robust accuracy (c) for increasing $s_\mathbf{x}$ scales in the purification (GAP) setup under APGD attack. Classifier: WRN-28-10. The shaded area shows the 95\% CI over four seeds.
    }
    \label{fig:scales}
\end{figure*}

\subsection{Quantitative Results}
\begin{figure*}[t]
    \centering
\subfigure[Synthesis (GAS).]{
\begin{minipage}{0.37\linewidth}
    \includegraphics[width=\cifarwidth px]{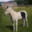}
    \includegraphics[width=\cifarwidth px]{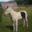}
    \includegraphics[width=\cifarwidth px]{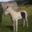}
    \includegraphics[width=\cifarwidth px]{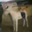}
    \includegraphics[width=\cifarwidth px]{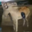}
    \\
    \includegraphics[width=\cifarwidth px]{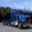}
    \includegraphics[width=\cifarwidth px]{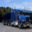}
    \includegraphics[width=\cifarwidth px]{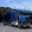}
    \includegraphics[width=\cifarwidth px]{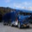}
    \includegraphics[width=\cifarwidth px]{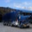}\\
    \includegraphics[width=\cifarwidth px] {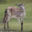}
    \includegraphics[width=\cifarwidth px]{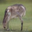}
    \includegraphics[width=\cifarwidth px]{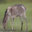}
    \includegraphics[width=\cifarwidth px]{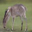}
    \includegraphics[width=\cifarwidth px]{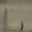}
    \begin{minipage}{0.188\linewidth}
        \scriptsize
        \centering
        $\smash{s_\mathbf{y}=0}$        
        \end{minipage} 
        \begin{minipage}{0.188\linewidth}
        \scriptsize
        \centering
        $\smash{s_\mathbf{y}=2^{-3}}$        
        \end{minipage} 
        \begin{minipage}{0.188\linewidth}
        \scriptsize
        \centering
        $\smash{s_\mathbf{y}=2^{-2}}$        
        \end{minipage} 
        \begin{minipage}{0.188\linewidth}
        \scriptsize
        \centering
        $\smash{s_\mathbf{y}=2^{-1}}$        
        \end{minipage} 
        \begin{minipage}{0.188\linewidth}
        \scriptsize
        \centering
        $\smash{s_\mathbf{y}=2^{0}}$        
        \end{minipage} \\
        \vspace{-1em}
\end{minipage}\label{fig:cifar_synthesis}}
\hspace{0.2cm}
\subfigure[Transform (GAT).]{
 \begin{minipage}{0.37\linewidth}
    \includegraphics[width=\cifarwidth px]{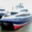}
    \includegraphics[width=\cifarwidth px]{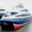}
        \includegraphics[width=\cifarwidth px]{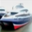}
        \includegraphics[width=\cifarwidth px]{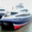}
        \includegraphics[width=\cifarwidth px]{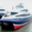} \\
    \includegraphics[width=\cifarwidth px]{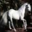}
        \includegraphics[width=\cifarwidth px]{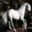}
        \includegraphics[width=\cifarwidth px]{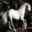}
        \includegraphics[width=\cifarwidth px]{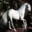}
        \includegraphics[width=\cifarwidth px]{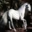} \\
    \includegraphics[width=\cifarwidth px]{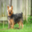}
        \includegraphics[width=\cifarwidth px]{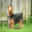}
        \includegraphics[width=\cifarwidth px]{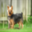}
        \includegraphics[width=\cifarwidth px]{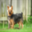}
        \includegraphics[width=\cifarwidth px]{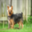} \\
        \begin{minipage}{0.188\linewidth}
        \scriptsize
        \centering
        \smash{Original}
        \end{minipage} 
        \begin{minipage}{0.188\linewidth}
        \scriptsize
        \centering
        \smash{$s_\mathbf{x}=32$}       
        \end{minipage} 
        \begin{minipage}{0.188\linewidth}
        \scriptsize
        \centering
        \smash{$s_\mathbf{x}=48$}       
        \end{minipage} 
        \begin{minipage}{0.188\linewidth}
        \scriptsize
        \centering
        \smash{$s_\mathbf{x}=64$}       
        \end{minipage} 
        \begin{minipage}{0.188\linewidth}
        \scriptsize
        \centering
        \smash{$s_\mathbf{x}=96$}       
        \end{minipage} \\
        \vspace{-1em}
\end{minipage}\label{fig:cifar_transformation}}
\vspace{-1em}
\caption{Examples on the CIFAR10 dataset. Fig.~\ref{fig:cifar_synthesis} shows the synthesis (GAS) setup and generates images of the classes ``horse", ``truck", and ``deer", which are classified as ``automobile", ``ship", and ``horse", respectively, as $s_\mathbf{y}$ increases. Fig.~\ref{fig:cifar_transformation} shows the transformation (GAT) setup and transforms images of the classes ``ship", ``horse", and ``dog", into adversarial examples classified as ``ship", ``deer", and ``cat". For $s_\mathbf{x}=32$, the images are outside of common perturbation norms\change{, i.e., $\ell_2=0.5$ and $\ell_\infty=8/255$}, but preserve image semantics. We show examples of selected baselines in Fig.~\ref{fig:cifar_baselines}.}
\vspace{-1.5em}
\label{fig:cifar_main}
\end{figure*}

\textbf{Evaluating Generative Adversarial Synthesis.}
As explained in Sec.~\ref{sec:synthesize}, \oursacro{} is capable of synthesizing adversarial examples. 
Fig.~\ref{fig:scale_left} and Fig.~\ref{fig:scale_middle} show the accuracy and the FID of a WRN-28-10 classifier as $s_\mathbf{y}$ increases, respectively. Notably, the classifier yields nearly identical performance as on real data when $s_\mathbf{y}=0$. However, even a minor increase of $s_\mathbf{y}$ to $0.125$ results in a substantial reduction in accuracy while maintaining a low FID. Setting $s_\mathbf{y}$ to 1.0 causes the classifier's performance to drop below random guessing levels for the CIFAR10 dataset. Additionally, Fig.~\ref{fig:cifar_synthesis} presents sample images generated at various scales. Notably, increasing $s_\mathbf{y}$ leads to subtle modifications in the images. Rather than introducing random noise, these changes maintain image coherence up to a scale of $s_\mathbf{y}=0.5$. Beyond this point, specifically at $s_\mathbf{y}=1.0$, there is a noticeable decline in image quality, as reflected by the FID.

Since our approach leverages a generative model, it enables the synthesis of an unlimited number of adversarial examples, thereby providing a more comprehensive robustness assessment. Moreover, in scenarios requiring the generation of adversarial examples, our method allows for rejection sampling at low $s_\mathbf{y}$ scales, ensuring the preservation of image quality. This is particularly important for adversarial training, where synthetic images can enhance robustness \citep{wang2023better}.

\textbf{Evaluating Generative Adversarial Transformation.}
Beyond the synthesis of new adversarial examples, our framework allows converting pre-existing images into adversarial ones as described in Sec.~\ref{sec:transform}. We show the accuracies and LPIPS scores of various attacks in Tab.~\ref{tab:generation}. Notably, \oursacro{} consistently achieves 0\% accuracy, lower than the $\ell_2$ and $\ell_0$ restricted methods across all three datasets, making it competitive to APGDT and LPA. This demonstrates \oursacro{}'s capability of generating adversarial examples. Surprisingly, the other unrestricted diffusion-based method, DiffAttack, yields considerably lower attack success rates. We attribute this discrepancy to the fact that it only leverages the last few iterations of the denoising diffusion process. Finally, we observe that the LPIPS scores of \oursacro{} are comparable to the restricted methods, and competitive to the minimum perturbation method FAB when applying additional LPIPS guidance (\oursacro{}-LPIPS), demonstrating its semantic preserving property. We present results on more classifiers in Tab.~\ref{tab:generation_additional} and on a high-resolution dataset in Tab.~\ref{tab:high_resolution} in the appendix. In Tab.~\ref{tab:purification}, we show the accuracies of APGD, APGDT, and \oursacro{} on robust classifiers. Notably, \oursacro{} demonstrates a considerably superior attack success rate compared to the PGD-based attacks. We attribute this to the more comprehensive robustness assessment of \oursacro{}. While most baselines only assess the robustness of adversarial examples on the $\ell_p$-constraint border, \oursacro{} draws samples from the distribution of semantics-preserving adversarial examples (see Sec.~\ref{sec:transform}).
\begin{table*}[t]
\caption{Robust accuracy and LPIPS scores for various attacks on CIFAR10, CIFAR100, and TinyImagenet. Best scores are in \textbf{bold}, second best \underline{underlined}.}
\label{tab:generation}
\centering
\scriptsize
\setlength{\tabcolsep}{8pt}
\arrayrulecolor{black!70}
\resizebox{\textwidth
}{!}{
\begin{tabular}{l c c c c c c} %
    \arrayrulecolor{black}
       &\multicolumn{3}{c}{Robust Accuracy in \% (${{\downarrow}}$)} &\multicolumn{3}{c}{LPIPS (${{\downarrow}}$)} \\
     \cmidrule(r){2-4}  \cmidrule(r){5-7}
    Dataset & CIFAR10 & CIFAR100 & TinyImagenet & CIFAR10 & CIFAR100 & TinyImagenet \\
    \midrule
   \multicolumn{1}{l}{\cellcolor{Gray}\textbf{$\mathbf{\ell_\infty}$ restricted}} & \multicolumn{6}{c}{\cellcolor{Gray}}\\
     \hspace{3mm}FGSM~\citep{goodfellow2014explaining} &  31.47{\tiny$\pm$13.39} & 10.82{\tiny$\pm$1.62} & 1.42{\tiny$\pm$0.17} & 30.27{\tiny$\pm$1.41} & 39.44{\tiny$\pm$1.45} & 180.76{\tiny$\pm$2.27}  \\
     \hspace{3mm}DI-FGSM~\citep{xie2019improving} & 0.54{\tiny$\pm$0.54} & 0.13{\tiny$\pm$0.10} & 0.04{\tiny$\pm$0.02} & 18.98{\tiny$\pm$3.63} & 22.87{\tiny$\pm$2.48} &125.46{\tiny$\pm$3.75}\\
     \hspace{3mm}SI-NI-FGSM~\citep{lin2019nesterov} & 3.01{\tiny$\pm$0.93} & 1.20{\tiny$\pm$0.16} & 0.69{\tiny$\pm$0.11} & 29.57{\tiny$\pm$3.69} & 40.92{\tiny$\pm$4.84} &  156.27{\tiny$\pm$3.49}\\
     \hspace{3mm}APGD~\citep{croce2020reliable}  &  0.18{\tiny$\pm$0.21}  &  \underline{0.10{\tiny$\pm$0.03}}  & 0.18{\tiny$\pm$0.03} & 12.40{\tiny$\pm$1.64} & 12.52{\tiny$\pm$1.52} & 88.10{\tiny$\pm$2.20}\\
     \hspace{3mm}APGDT~\citep{croce2020reliable}  &  \textbf{0.00{\tiny$\pm$0.00}}  &  \textbf{0.00{\tiny$\pm$0.00}}  & \textbf{0.00{\tiny$\pm$0.00}} & 12.17{\tiny$\pm$0.11} & 11.56{\tiny$\pm$0.53} & 65.64{\tiny$\pm$1.55}\\
     \hspace{3mm}Square~\citep{andriushchenko2020square} & 0.25{\tiny$\pm$0.24} & 0.19{\tiny$\pm$0.04} & 0.51{\tiny$\pm$0.05} & 126.20{\tiny$\pm$1.61} & 88.43{\tiny$\pm$0.9} & 127.93{\tiny$\pm$1.46}\\
     \hspace{3mm}FAB~\citep{croce2020minimally} & 1.67{\tiny$\pm$1.56} & 0.76{\tiny$\pm$0.06} & 0.11{\tiny$\pm$0.19} & 0.78{\tiny$\pm$0.36} & \underline{0.15{\tiny$\pm$0.01}} & \underline{7.05{\tiny$\pm$0.09}}\\ 
   \multicolumn{1}{l}{\cellcolor{Gray}\textbf{$\mathbf{\ell_2}$ restricted}} & \multicolumn{6}{c}{\cellcolor{Gray}}\\
     \hspace{3mm}APGD~\citep{croce2020reliable} &  1.21{\tiny$\pm$0.05}  & 0.69{\tiny$\pm$0.01} & 0.15{\tiny$\pm$0.05} & 2.48{\tiny$\pm$0.18} & 2.66{\tiny$\pm$0.18} & 96.01{\tiny$\pm$2.92}\\
     \hspace{3mm}APGDT~\citep{croce2020reliable} &  0.11{\tiny$\pm$0.01} & 0.09{\tiny$\pm$0.01} & \textbf{0.00{\tiny$\pm$0.00}} & 2.51{\tiny$\pm$0.1} & 2.33{\tiny$\pm$0.17} & 101.13{\tiny$\pm$2.62}\\
 \hspace{3mm}Square~\citep{andriushchenko2020square} & 19.67{\tiny$\pm$0.27} & 7.02{\tiny$\pm$0.42} & 1.26{\tiny$\pm$0.10} & 8.54{\tiny$\pm$0.13} & 10.06{\tiny$\pm$0.58} & 151.46{\tiny$\pm$2.58}\\
 \hspace{3mm}FAB~\citep{croce2020minimally} & 7.41{\tiny$\pm$6.19} & 1.44{\tiny$\pm$0.33} & \underline{0.01{\tiny$\pm$0.01}} & \textbf{0.36{\tiny$\pm$0.06}} & \textbf{0.10{\tiny$\pm$0.01}} & \textbf{0.56{\tiny$\pm$0.06}} \\ 
   \multicolumn{1}{l}{\cellcolor{Gray}\textbf{$\mathbf{\ell_0}$ restricted}} & \multicolumn{6}{c}{\cellcolor{Gray}}\\    \hspace{3mm}OnePixel~\citep{su2019one} & 82.82{\tiny$\pm$0.94} & 59.17{\tiny$\pm$0.77} & 59.42{\tiny$\pm$0.38} & 10.67{\tiny$\pm$1.16} & 13.65{\tiny$\pm$0.54} & 11.08{\tiny$\pm$0.10}\\
   \multicolumn{1}{l}{\cellcolor{Gray}\textbf{Unrestricted}} & \multicolumn{6}{c}{\cellcolor{Gray}}\\    %
   \hspace{3mm}CAA~\citep{hsiung2022caa} & 43.23{\tiny$\pm$0.71} & 12.88{\tiny$\pm$0.41}  & 8.81{\tiny$\pm$0.44} & 1564.62{\tiny$\pm$23.23} & 1266.86{\tiny$\pm$22.07} & 879.52{\tiny$\pm$14.94}\\
   \hspace{3mm}PPGD~\citep{laidlaw2020perceptual} & 31.82{\tiny$\pm$2.77} & 39.76{\tiny$\pm$2.08} & 2.76{\tiny$\pm$0.10} & 10.70{\tiny$\pm$0.12} & 7.23{\tiny$\pm$0.18} & 31.13{\tiny$\pm$0.63}\\
    \hspace{3mm}LPA~\citep{laidlaw2020perceptual}& \underline{0.04{\tiny$\pm$0.05}}& \textbf{0.00{\tiny$\pm$0.00}} & \textbf{0.00{\tiny$\pm$0.00}} &  25.41{\tiny$\pm$6.50} & 40.08{\tiny$\pm$9.57} & 339.48{\tiny$\pm$6.03}\\
\hspace{3mm}DiffAttack~\citep{chen2023diffusion} & 14.40{\tiny$\pm$0.97} & 4.89{\tiny$\pm$1.57} & 2.13{\tiny$\pm$0.09} & 637.89{\tiny$\pm$3.68} & 626.99{\tiny$\pm$4.98} & 808.90{\tiny$\pm$6.36}  \\
\hspace{3mm}\oursacro{} (Ours) & \textbf{0.00{\tiny$\pm$0.00}} & \textbf{0.00{\tiny$\pm$0.00}} & \textbf{0.00{\tiny$\pm$0.00}} & 4.39{\tiny$\pm$0.13} & 4.28{\tiny$\pm$0.22} & 109.11{\tiny$\pm$0.55} \\
\hspace{3mm}\oursacro{}-LPIPS (Ours) & \textbf{0.00{\tiny$\pm$0.00}} & 0.01{\tiny$\pm$0.01} & \textbf{0.00{\tiny$\pm$0.00}} & \underline{0.63{\tiny$\pm$0.03}} & 0.54{\tiny$\pm$0.02} & 42.83{\tiny$\pm$3.39} \\
    \bottomrule
\end{tabular}}
\vspace{-2.25em}
\end{table*}

\textbf{Evaluating Generative Adversarial Purification.}
Finally, we examine the purification ability of \oursacro{}. Tab.~\ref{tab:purification} shows the purification results for various methods on the CIFAR10 dataset. Our results show that \oursacro{} consistently achieves state-of-the-art performance in robust accuracy, outperforming other adversarial purification and training methods. Notably, \oursacro{} not only successfully defends attacks but also maintains a high level of clean accuracy comparable to that of adversarial training. This demonstrates \oursacro{}'s capability to preserve the core semantics while effectively neutralizing the impact of adversarial perturbations by projecting samples on the manifold of clean images. Overall, our results indicate that purification methods can defend better against adversarial attacks than adversarial training approaches, which we attribute to the preprocessor-blackbox setting. Note that it is not possible to detect adversarial examples. Therefore, the purification needs to be applied to all images. However, \oursacro{} still achieves a high clean accuracy. We demonstrate its applicability to common corruptions in App.~\ref{sec:corruptions}.
\begin{table*}[t!]
\caption{CIFAR10 robust accuracy of different adversarial training and purification methods for the attacks APGD, APGDT, and \oursacro{}. If multiple threat models exist, we denote results as $\ell_\infty$/$\ell_2$. Best purification scores are in \textbf{bold}, best attack success rates are \underline{underlined}.}
\vspace{-0.5em}
\label{tab:purification}
\scriptsize
\begin{center}
  \resizebox{\linewidth}{!}{
  \begin{tabular}{l c c c c c c c}
    \multirow{2}{*}{Model} & \multirow{ 2}{*}{Clean Accuracy} & \multicolumn{2}{c}{APGD} & \multicolumn{2}{c}{APGDT} & \oursacro{}-GAT (Ours) & \multirow{2}{*}{Architecture} \\
    & & $\ell_\infty$ & $\ell_2$ & $\ell_\infty$ & $\ell_2$ & Unrestricted\\
    \cmidrule(r){1-8}
   \multicolumn{1}{l}{\cellcolor{Gray}\textbf{Adversarial Training}} & \multicolumn{7}{c}{\cellcolor{Gray}}\\\hspace{3mm}\citep{cui2023decoupled} & 92.16 & 70.36 & - & 68.43 & - & \underline{47.69} & WRN-28-10\\\hspace{3mm}\citep{wang2023better} & 92.44 / 95.16 & 70.08 & 84.52 & 68.04 & 83.88 & \underline{45.33 / 38.49} & WRN-28-10\\\hspace{3mm}\citep{wang2023better} & 93.25 / 95.54 & 73.29 & 85.65 & 71.42 & 85.28 & \underline{41.52 / 41.37} & WRN-70-16\\
   \hspace{3mm}\citep{peng2023robust} & 93.27 & 73.67 & - & 71.82 & - & \underline{38.87} & RaWRN-70-16\\
    \multicolumn{1}{l}{\cellcolor{Gray}\textbf{Adversarial Purification}} & \multicolumn{7}{c}{\cellcolor{Gray}}\\
      \hspace{3mm}ADP~\citep{yoon2021adversarial} & 93.09 & - & - & 85.45 & - & - & WRN-28-10\\
      \hspace{3mm}DiffPure~\citep{nie2022DiffPure} & 89.02 & \underline{87.72} & 88.46 & 88.30 & 88.18 & 88.57 & WRN-28-10\\ %
            \hspace{3mm}\oursacro{}-GAP (Ours) & 93.93$\pm$0.12 & \textbf{91.34$\pm$0.46} & \textbf{92.13$\pm$1.41} & \textbf{90.25$\pm$0.44} & \textbf{90.89$\pm$0.40} & \textbf{90.74$\pm$0.67} & WRN-28-10\\
      \bottomrule
  \end{tabular}}
\end{center}
\vspace{-2.5em}
\end{table*} 

\textbf{Hyperparameter study.}
We explore the impact of the scale parameters $s_\mathbf{y}$ and $s_\mathbf{x}$ on accuracy and FID, as depicted in Fig.~\ref{fig:scales}. 
In Fig.~\ref{fig:scale_right}, we examine the efficacy of purification against adversarial attacks of APGD under both $\ell_2$ and $\ell_\infty$ norms across different $s_\mathbf{x}$ scales. At $s_\mathbf{x}=0$, the generated images are unconditional without guidance and independent of the input. Therefore, the robust accuracy equals random guessing. As $s_\mathbf{x}$ increases, the accuracy improves, reaching a performance plateau at approximately $s_\mathbf{x}=10$. Increasing $s_\mathbf{x}$ further reduces the accuracy as the sampled images start to resemble adversarial perturbations. In practice, we scale $s$ by $t^{-1}$.

\begin{wraptable}[14]{r}{0.45\textwidth}
    \vspace{-1em}
  \caption{Robust accuracy and median $\ell_2$ distances for various hyperparameter configurations. Best scores are in \textbf{bold}.}
  \vspace{-0.75em}
  \label{tab:hyperparameter}
  \begin{center}
  \scriptsize
\resizebox{0.45\textwidth}{!}{%
  \begin{tabular}{@{}l c c c c@{}}
   &\multicolumn{2}{c}{Robust Accuracy in \% (${{\downarrow}}$)} &\multicolumn{2}{c}{Median $\ell_2$ distance} \\
     \cmidrule(r){2-3}  \cmidrule(r){4-5}
    Dataset & CIFAR10 & CIFAR100 &  CIFAR10 & CIFAR100   \\
        \midrule
    \multicolumn{1}{l}{\cellcolor{Gray}$\mathbf{s_\mathbf{y}=48}$} & \multicolumn{4}{c}{\cellcolor{Gray}}\\
    \hspace{4mm}$s_\mathbf{x}=16$ & \textbf{0.10} & \textbf{0.02} & 1.12 & 1.09 \\
    \hspace{4mm}$s_\mathbf{x}=32$ & 0.23 & \textbf{0.02} & 0.65 & 0.64 \\
    \hspace{4mm}$s_\mathbf{x}=48$ & 0.32 & 0.03 & 0.49 & 0.49 \\
    \hspace{4mm}$s_\mathbf{x}=64$ & 0.34 & 0.04 & 0.43 &  0.40\\
    \multicolumn{1}{l}{\cellcolor{Gray}$\mathbf{s_\mathbf{y}=64}$} & \multicolumn{4}{c}{\cellcolor{Gray}}\\
    \hspace{4mm}$s_\mathbf{x}=48$ & 0.22 & 0.17 & 0.50 & 0.49
\\
    \hspace{4mm}$s_\mathbf{x}=64$ & 0.24 & \textbf{0.02} & 0.43 & 0.40\\
    \hspace{4mm}$s_\mathbf{x}=96$ & 0.28 & 0.03 & 0.35 & 0.30\\
    \multicolumn{1}{l}{\cellcolor{Gray}$\mathbf{s_\mathbf{y}=96}$} & \multicolumn{4}{c}{\cellcolor{Gray}}\\
    \hspace{4mm}$s_\mathbf{x}=48$ & \textbf{0.10} & 0.17 & 0.51 & 0.50 \\
    \hspace{4mm}$s_\mathbf{x}=64$ & 0.11 & 0.21 & 0.44 & 0.40 \\
    \hspace{4mm}$s_\mathbf{x}=96$ & 0.13 & 0.34 & 0.35 & 0.30\\
    \bottomrule
\end{tabular}}
\vspace{-3.5em}
  \end{center}
\end{wraptable}

Finally, Tab.~\ref{tab:hyperparameter} shows the robust accuracy and median $\ell_2$ distances across different scale configurations for the CIFAR10 and CIFAR100 datasets. We can observe that an increase in $s_\mathbf{y}$ leads to reduced classifier accuracy for CIFAR10, improving the efficacy of the adversarial attacks. A rise in $s_\mathbf{x}$, however, increases the accuracy as the generated image closer resembles the original. The median $\ell_2$ distance exhibits a similar behavior. While a lower $s_\mathbf{y}$ yields no difference for both datasets, increasing $s_\mathbf{x}$ decreases the median distances for CIFAR10 and CIFAR100. In Fig.~\ref{fig:cifar_transformation}, we show examples across various $s_\mathbf{x}$ scales on the CIFAR10 dataset. Notably, all scales preserve the image semantics and do not display any observable differences. In practice, we iteratively increase the scale $s_\mathbf{y}$ if the attack is not successful.

\subsection{Qualitative Analysis}
To investigate the quality of the adversarial attacks, we deploy \oursacro{} on the ImageNet dataset \citep{deng2009imagenet} with a resolution of $256\times 256$. We use the latent diffusion model DiT proposed by \citet{Peebles2022DiT}, along with a pre-trained latent classifier from \citep{kim2022refining}. The images are sampled using the denoising procedure by \citet{kollovieh2023predict} as explained in Sec.~\ref{sec:synthesize}. Note that as the generative process is performed in the latent space, the model has more freedom in terms of reconstruction.

We show an example image of a tiger shark in Fig.~\ref{fig:main} with corresponding adversarial attacks. While the classifier correctly identifies the tiger shark in the baseline image, it fails to do so in the generated adversarial examples. Notably, the $\ell_p$-bounded methods display noticeable noisy fragments. In contrast, \oursacro{} produces clean adversarial examples, altering only minor details while retaining the core semantics --- most notably, the removal of a small fish --- which prove to be important classification cues. We provide further examples for GAS in Sec.~\ref{sec:GAS_images} and for GAT in Sec.~\ref{sec:GAA_images}. The synthetic images display a high degree of realism, and the transformed images show visible differences while preserving the semantics of the original image.

\subsection{Human Study}

To evaluate whether \oursacro{} generates semantics-preserving adversarial examples, we perform a human study on adversarially modified (real) as well as synthetically generated images. 
For the study, we choose CIFAR10 images as it (1) avoids any class-selection bias, whereas high-resolution datasets usually contain many classes only \change{distinguishable} by human experts; and (2) is the most commonly used dataset in related work. Hyperparameters are set to produce an interesting regime, where the generated adversarial images are significantly outside common $\ell_p$-norm balls and constitute strong attacks for the classifier in question.
In particular, we randomly sample five images from each class to generate 50 adversarial examples using $s_\mathbf{x}=16$ and $s_\mathbf{y}=48$. These adversarial examples have an average $\ell_2$-norm difference to their clean counterparts of $0.68\pm0.24$, exceeding the common $\ell_2$-norm ball constraint of 0.5 \citep{croce2020robustbench} by on average 36\%. For the synthetic examples, we generate 50 images without ($s_\mathbf{y}=0$) and 50 images with guidance $(s_\mathbf{y}=0.125)$, again in a class-balanced fashion. For the adversarial guided synthetic examples, we employ rejection sampling to only consider images that lead to misclassification by the classifier. 
To ensure high data quality for the study, we used the Prolific platform \citep{eyal2021data} to employ 60 randomly chosen human evaluators to label the 200 images.
To avoid bias, we presented the adversarial examples (synthetic or modified) before the unperturbed examples and introduced the category "Other / I don't know".
\begin{wraptable}[11]{r}{0.422\textwidth}
    \vspace{-0.5em}
    \caption{Human study to evaluate the adversarial examples of \oursacro{}. The human ACC corresponds to the majority vote.} %
  \label{tab:humanstudy}
  \begin{center}
  \scriptsize
    \vspace{-0.2em}
\resizebox{0.4\textwidth}{!}{%
  \begin{tabular}{@{}l r r @{}}
    Dataset & Model ACC & Human ACC  \\
    \midrule
    \multicolumn{1}{l}{\cellcolor{Gray}\textbf{Clean}} & \multicolumn{2}{c}{\cellcolor{Gray}}\\
    \hspace{4mm}Real      & $98\%$ & $100\%$ \\
    \hspace{4mm}Synthetic & $94\%$ & $94\%$\\
    \multicolumn{1}{l}{\cellcolor{Gray}\textbf{Adversarial}} & \multicolumn{2}{c}{\cellcolor{Gray}}\\
     \hspace{4mm}Real      & $2\%$ & $94\%$  \\
     \hspace{4mm}Synthetic & $0\%$ & $70\%$ \\
    \bottomrule
\end{tabular}}
\vspace{-3.25em}
  \end{center}
\end{wraptable}

We compute human accuracy by choosing the majority vote class of all 60 human evaluators and compare it with the ground truth class. We show the results of the human study in Tab.~\ref{tab:humanstudy}. %
Notably, humans can still accurately classify 94\% of the adversarial modified images despite significantly larger $\ell_2$ distances, establishing \emph{almost perfect} semantic preservation for GAT. For GAS, humans classify 70\% of the (successful) synthetic adversarial images correctly. This is lower than for adversarial modification and shows that the generation of completely synthetic semantics-preserving adversarial examples is a harder task than adversarial modification. Still, GAS achieves \emph{good} semantic preservation, significantly outperforming random guessing (10\%). We believe it is critical that semantic preservation of unrestricted attacks is evaluated through human studies as done in some early works \citep{song2018unrestricted, khoshpasand2020unrestricted}. As this is missing in all related unrestricted attack works used as baselines in this work, we hope to contribute to establishing this as an evaluation standard, and that our results can serve as interesting baselines for future works.

\section{Related Work}
\textbf{Diffusion Models.}
Diffusion models \citep{sohl2015deep, ho2020denoising} and score-based generative models \citep{song2020score} received significant attention in recent years, owing to their remarkable performance across various domains \citep{kong2020diffwave, lienen2023generative, kollovieh2023predict} and have since emerged as the go-to methodology for many generative tasks. \citet{dhariwal2021diffusion} proposed diffusion guidance to perform conditional sampling using unconditional models. 
A recent study has shown that classifiers can enhance their robust accuracy when training on images generated by diffusion models \citep{wang2023better}, demonstrating the usefulness and potential of diffusion models in the robustness domain. %

\textbf{Adversarial Attacks.}
An important line of work are white-box approaches, which have full access to the model parameters and gradients, such as the fast gradient sign method (FGSM) introduced by \citet{goodfellow2014explaining}. While FGSM and its subsequent extensions \citep{xie2019improving, dong2018boosting, lin2019nesterov, wang2021adversarial} primarily focus on perturbations constrained by the $\ell_\infty$ norm, other white-box techniques employ projected gradient descent and explore a broader range of perturbation norms \citep{madry2017towards, zhang2019theoretically}. In contrast, black-box attacks are closer to real-world scenarios and do not have access to model parameters or gradients \citep{narodytska2016simple, brendel2017decision, andriushchenko2020square}. \change{As \oursacro{}-GAT and \oursacro{}-GAS rely on the gradients of the classifier to compute guidance scores, they are categorized as white-box attacks.}

\textbf{Diffusion-Based Attacks.} Two recent works by \citet{chen2023diffusion} and \citet{xue2023diffusion} propose DiffAttack and Diff-PGD, respectively. Diff-PGD performs projected gradient descent in the latent diffusion space to obtain $\ell_\infty$-bounded adversarial examples, whereas DiffAttack generates \emph{unrestricted} adversarial examples by leveraging a latent diffusion model. 
However, as both methods employ only the final denoising stages of the diffusion process in a similar fashion to SDEdit \citep{meng2021sdedit}, the adversarial perturbations only incorporate changes of high-level features. Finally, \citet{chen2023content} implement PGD in the $\ell_\infty$-norm within the latent space of stable diffusion. In parallel, \citet{chen2023advdiffuser} apply PGD iteratively at each step of the diffusion process and combine it with adversarial inpainting. %
Unlike previous works, \oursacro{} does not rely on PGD in the latent space for its attack and semantic preservation, but solely leverages the diffusion manifold in combination with a task-specific guidance.

\textbf{Adversarial Purification.}
In response to the introduction of adversarial attacks, a variety of adversarial purification methods to defend machine learning models have emerged. 
Early works utilized Generative Adversarial Networks (GANs) \citet{song2017pixeldefend, song2018unrestricted, samangouei2018defense} and Energy-Based Models (EBMs) \citep{hill2020stochastic} to remove adversarial perturbations from images.
More recent methods have shifted focus towards score-based generative models, like ADP \citep{yoon2021adversarial}, and diffusion models, such as DiffPure~\citep{nie2022DiffPure}. However, ADP and DiffPure only denoise with small noise magnitudes during the purification process and are thereby limited to correcting high-level adversarial features, whereas \oursacro{} traverses the whole diffusion process,  providing more flexibility in purifying perturbations. \citet{kang2023diffattack} \change{have} recently shown that these purification methods decrease in effectiveness in a white-box setting by evasion attacks. However, as previously mentioned, we focus on preprocessor black-box attacks, which are more relevant in real-world problems.

\section{Discussion}\label{sec:discussion}
\textbf{Limitations and Future Work.}
Our work demonstrates the potential and capabilities of score-based generative models in the realm of adversarial attacks and robustness. While \oursacro{} is able to generate and purify adversarial attacks, some drawbacks remain. Primarily, the evaluation of unrestricted attacks remains challenging. We resolve this limitation by performing a human study and argue that this should become standard. Moreover, the proposed purification approach is only applicable to a preprocessor-blackbox setting, as computing the gradients of the generative process efficiently is an open problem. %

\textbf{Conclusion.} In this work, we address the question of how to generate \emph{unrestricted} adversarial examples. We introduce \oursacro{}, a novel framework that bridges the gap between adversarial attacks and score-based generative models. Utilizing diffusion guidance and pre-trained models, \oursacro{} can synthesize new adversarial attacks, transform existing images into adversarial examples, and purify images, thereby enhancing the empirical robust accuracy of classifiers. Our results indicate that \oursacro{} can effectively generate semantics-preserving adversarial images beyond the limitations of the $\ell_p$-norms. Our experimental evaluation demonstrates that \oursacro{} matches the performance of existing state-of-the-art attacks and defenses. We see unrestricted adversarial examples - as generated by our work - as vital to achieve a holistic view of robustness and complementary to hand-picked common corruptions \citep{kar20223d} or classical $\ell_p$ threat models.  
\paragraph{Broader Impact}
This work contributes to the domain of robustness, focusing on unrestricted adversarial attacks. Our framework, \oursacro{}, is designed for the generation and purification of adversarial images. While there exists the potential for malicious misuse, we hope for our insights to enhance the understanding of machine learning models' robustness. Moreover, despite the competitive empirical performance of \oursacro{}, we advise against relying solely on the algorithm.
\section*{Acknowledgments}
This paper has been supported by the Munich Center for Machine Learning and by the DAAD program Konrad Zuse Schools of Excellence in Artificial Intelligence, sponsored by the German Federal Ministry of Education and Research, and the German Research Foundation, grant GU 1409/4-1.
\bibliography{main}

\begin{thebibliography}{60}
\providecommand{\natexlab}[1]{#1}
\providecommand{\url}[1]{\texttt{#1}}
\expandafter\ifx\csname urlstyle\endcsname\relax
  \providecommand{\doi}[1]{doi: #1}\else
  \providecommand{\doi}{doi: \begingroup \urlstyle{rm}\Url}\fi

\bibitem[Andriushchenko et~al.(2020)Andriushchenko, Croce, Flammarion, and
  Hein]{andriushchenko2020square}
Maksym Andriushchenko, Francesco Croce, Nicolas Flammarion, and Matthias Hein.
\newblock Square attack: a query-efficient black-box adversarial attack via
  random search.
\newblock In \emph{European conference on computer vision}, pp.\  484--501.
  Springer, 2020.

\bibitem[Brendel et~al.(2017)Brendel, Rauber, and Bethge]{brendel2017decision}
Wieland Brendel, Jonas Rauber, and Matthias Bethge.
\newblock Decision-based adversarial attacks: Reliable attacks against
  black-box machine learning models.
\newblock \emph{arXiv preprint arXiv:1712.04248}, 2017.

\bibitem[Chen et~al.(2023{\natexlab{a}})Chen, Chen, Chen, Zhang, Zou, and
  Shi]{chen2023diffusion}
Jianqi Chen, Hao Chen, Keyan Chen, Yilan Zhang, Zhengxia Zou, and Zhenwei Shi.
\newblock Diffusion models for imperceptible and transferable adversarial
  attack.
\newblock \emph{arXiv preprint arXiv:2305.08192}, 2023{\natexlab{a}}.

\bibitem[Chen et~al.(2023{\natexlab{b}})Chen, Gao, Zhao, Ye, and
  Xu]{chen2023advdiffuser}
Xinquan Chen, Xitong Gao, Juanjuan Zhao, Kejiang Ye, and Cheng-Zhong Xu.
\newblock Advdiffuser: Natural adversarial example synthesis with diffusion
  models.
\newblock In \emph{Proceedings of the IEEE/CVF International Conference on
  Computer Vision}, pp.\  4562--4572, 2023{\natexlab{b}}.

\bibitem[Chen et~al.(2023{\natexlab{c}})Chen, Li, Wu, Jiang, Ding, and
  Zhang]{chen2023content}
Zhaoyu Chen, Bo~Li, Shuang Wu, Kaixun Jiang, Shouhong Ding, and Wenqiang Zhang.
\newblock Content-based unrestricted adversarial attack.
\newblock \emph{arXiv preprint arXiv:2305.10665}, 2023{\natexlab{c}}.

\bibitem[Croce \& Hein(2020{\natexlab{a}})Croce and Hein]{croce2020minimally}
Francesco Croce and Matthias Hein.
\newblock Minimally distorted adversarial examples with a fast adaptive
  boundary attack.
\newblock In \emph{International Conference on Machine Learning}, pp.\
  2196--2205. PMLR, 2020{\natexlab{a}}.

\bibitem[Croce \& Hein(2020{\natexlab{b}})Croce and Hein]{croce2020reliable}
Francesco Croce and Matthias Hein.
\newblock Reliable evaluation of adversarial robustness with an ensemble of
  diverse parameter-free attacks.
\newblock In \emph{ICML}, 2020{\natexlab{b}}.

\bibitem[Croce et~al.(2020)Croce, Andriushchenko, Sehwag, Debenedetti,
  Flammarion, Chiang, Mittal, and Hein]{croce2020robustbench}
Francesco Croce, Maksym Andriushchenko, Vikash Sehwag, Edoardo Debenedetti,
  Nicolas Flammarion, Mung Chiang, Prateek Mittal, and Matthias Hein.
\newblock Robustbench: a standardized adversarial robustness benchmark.
\newblock \emph{arXiv preprint arXiv:2010.09670}, 2020.

\bibitem[Cui et~al.(2023)Cui, Tian, Zhong, Qi, Yu, and Zhang]{cui2023decoupled}
Jiequan Cui, Zhuotao Tian, Zhisheng Zhong, Xiaojuan Qi, Bei Yu, and Hanwang
  Zhang.
\newblock Decoupled kullback-leibler divergence loss.
\newblock \emph{arXiv preprint arXiv:2305.13948}, 2023.

\bibitem[Deng et~al.(2009)Deng, Dong, Socher, Li, Li, and
  Fei-Fei]{deng2009imagenet}
Jia Deng, Wei Dong, Richard Socher, Li-Jia Li, Kai Li, and Li~Fei-Fei.
\newblock Imagenet: A large-scale hierarchical image database.
\newblock In \emph{2009 IEEE conference on computer vision and pattern
  recognition}, pp.\  248--255. Ieee, 2009.

\bibitem[Dhariwal \& Nichol(2021)Dhariwal and Nichol]{dhariwal2021diffusion}
Prafulla Dhariwal and Alexander Nichol.
\newblock Diffusion models beat gans on image synthesis.
\newblock \emph{Advances in neural information processing systems},
  34:\penalty0 8780--8794, 2021.

\bibitem[Dong et~al.(2023)Dong, Chen, Xie, Lai, and Chen]{dong2023adversarial}
Junhao Dong, Junxi Chen, Xiaohua Xie, Jianhuang Lai, and Hao Chen.
\newblock Adversarial attack and defense for medical image analysis: Methods
  and applications.
\newblock \emph{arXiv preprint arXiv:2303.14133}, 2023.

\bibitem[Dong et~al.(2018)Dong, Liao, Pang, Su, Zhu, Hu, and
  Li]{dong2018boosting}
Yinpeng Dong, Fangzhou Liao, Tianyu Pang, Hang Su, Jun Zhu, Xiaolin Hu, and
  Jianguo Li.
\newblock Boosting adversarial attacks with momentum.
\newblock In \emph{Proceedings of the IEEE conference on computer vision and
  pattern recognition}, pp.\  9185--9193, 2018.

\bibitem[Eyal et~al.(2021)Eyal, David, Andrew, Zak, and
  Ekaterina]{eyal2021data}
Peer Eyal, Rothschild David, Gordon Andrew, Evernden Zak, and Damer Ekaterina.
\newblock Data quality of platforms and panels for online behavioral research.
\newblock \emph{Behavior Research Methods}, pp.\  1--20, 2021.

\bibitem[Eykholt et~al.(2018)Eykholt, Evtimov, Fernandes, Li, Rahmati, Xiao,
  Prakash, Kohno, and Song]{eykholt2018physical}
Kevin Eykholt, Ivan Evtimov, Earlence Fernandes, Bo~Li, Amir Rahmati, Chaowei
  Xiao, Atul Prakash, Tadayoshi Kohno, and Dawn Song.
\newblock Robust physical-world attacks on deep learning visual classification.
\newblock In \emph{2018 IEEE/CVF Conference on Computer Vision and Pattern
  Recognition}, pp.\  1625--1634, 2018.
\newblock \doi{10.1109/CVPR.2018.00175}.

\bibitem[Fursov et~al.(2021)Fursov, Morozov, Kaploukhaya, Kovtun,
  Rivera-Castro, Gusev, Babaev, Kireev, Zaytsev, and
  Burnaev]{fursov2021adversarial}
Ivan Fursov, Matvey Morozov, Nina Kaploukhaya, Elizaveta Kovtun, Rodrigo
  Rivera-Castro, Gleb Gusev, Dmitry Babaev, Ivan Kireev, Alexey Zaytsev, and
  Evgeny Burnaev.
\newblock Adversarial attacks on deep models for financial transaction records.
\newblock In \emph{Proceedings of the 27th ACM SIGKDD Conference on Knowledge
  Discovery \& Data Mining}, pp.\  2868--2878, 2021.

\bibitem[Goodfellow et~al.(2014)Goodfellow, Shlens, and
  Szegedy]{goodfellow2014explaining}
Ian~J Goodfellow, Jonathon Shlens, and Christian Szegedy.
\newblock Explaining and harnessing adversarial examples.
\newblock \emph{arXiv preprint arXiv:1412.6572}, 2014.

\bibitem[Gosch et~al.(2023)Gosch, Sturm, Geisler, and
  G{\"u}nnemann]{gosch2023revisiting}
Lukas Gosch, Daniel Sturm, Simon Geisler, and Stephan G{\"u}nnemann.
\newblock Revisiting robustness in graph machine learning.
\newblock \emph{arXiv preprint arXiv:2305.00851}, 2023.

\bibitem[Hendrycks \& Dietterich(2019)Hendrycks and
  Dietterich]{hendrycks2018benchmarking}
Dan Hendrycks and Thomas Dietterich.
\newblock Benchmarking neural network robustness to common corruptions and
  perturbations.
\newblock In \emph{International Conference on Learning Representations}, 2019.

\bibitem[Hendrycks et~al.(2022)Hendrycks, Carlini, Schulman, and
  Steinhardt]{hendrycks2022unsolved}
Dan Hendrycks, Nicholas Carlini, John Schulman, and Jacob Steinhardt.
\newblock Unsolved problems in ml safety, 2022.

\bibitem[Heusel et~al.(2017)Heusel, Ramsauer, Unterthiner, Nessler, and
  Hochreiter]{heusel2017gans}
Martin Heusel, Hubert Ramsauer, Thomas Unterthiner, Bernhard Nessler, and Sepp
  Hochreiter.
\newblock Gans trained by a two time-scale update rule converge to a local nash
  equilibrium.
\newblock \emph{Advances in neural information processing systems}, 30, 2017.

\bibitem[Hill et~al.(2020)Hill, Mitchell, and Zhu]{hill2020stochastic}
Mitch Hill, Jonathan Mitchell, and Song-Chun Zhu.
\newblock Stochastic security: Adversarial defense using long-run dynamics of
  energy-based models.
\newblock \emph{arXiv preprint arXiv:2005.13525}, 2020.

\bibitem[Ho et~al.(2020)Ho, Jain, and Abbeel]{ho2020denoising}
Jonathan Ho, Ajay Jain, and Pieter Abbeel.
\newblock Denoising diffusion probabilistic models.
\newblock \emph{Advances in neural information processing systems},
  33:\penalty0 6840--6851, 2020.

\bibitem[Hsiung et~al.(2023)Hsiung, Tsai, Chen, and Ho]{hsiung2022caa}
Lei Hsiung, Yun-Yun Tsai, Pin-Yu Chen, and Tsung-Yi Ho.
\newblock {Towards Compositional Adversarial Robustness: Generalizing
  Adversarial Training to Composite Semantic Perturbations}.
\newblock In \emph{{IEEE/CVF} Conference on Computer Vision and Pattern
  Recognition, {CVPR}}. {IEEE}, June 2023.

\bibitem[Kang et~al.(2023)Kang, Song, and Li]{kang2023diffattack}
Mintong Kang, Dawn Song, and Bo~Li.
\newblock Diffattack: Evasion attacks against diffusion-based adversarial
  purification.
\newblock \emph{arXiv preprint arXiv:2311.16124}, 2023.

\bibitem[Kar et~al.(2022)Kar, Yeo, Atanov, and Zamir]{kar20223d}
O{\u{g}}uzhan~Fatih Kar, Teresa Yeo, Andrei Atanov, and Amir Zamir.
\newblock 3d common corruptions and data augmentation.
\newblock In \emph{Proceedings of the IEEE/CVF Conference on Computer Vision
  and Pattern Recognition}, pp.\  18963--18974, 2022.

\bibitem[Karras et~al.(2022)Karras, Aittala, Aila, and
  Laine]{karras2022elucidating}
Tero Karras, Miika Aittala, Timo Aila, and Samuli Laine.
\newblock Elucidating the design space of diffusion-based generative models.
\newblock \emph{Advances in Neural Information Processing Systems},
  35:\penalty0 26565--26577, 2022.

\bibitem[Khoshpasand \& Ghorbani(2020)Khoshpasand and
  Ghorbani]{khoshpasand2020unrestricted}
Mehrgan Khoshpasand and Ali Ghorbani.
\newblock On the generation of unrestricted adversarial examples.
\newblock In \emph{2020 50th Annual IEEE/IFIP International Conference on
  Dependable Systems and Networks Workshops (DSN-W)}, pp.\  9--15, 2020.

\bibitem[Kim et~al.(2022)Kim, Kim, Kang, and Moon]{kim2022refining}
Dongjun Kim, Yeongmin Kim, Wanmo Kang, and Il-Chul Moon.
\newblock Refining generative process with discriminator guidance in
  score-based diffusion models.
\newblock \emph{arXiv preprint arXiv:2211.17091}, 2022.

\bibitem[Kim(2020)]{kim2020torchattacks}
Hoki Kim.
\newblock Torchattacks: A pytorch repository for adversarial attacks.
\newblock \emph{arXiv preprint arXiv:2010.01950}, 2020.

\bibitem[Kollovieh et~al.(2023)Kollovieh, Ansari, Bohlke-Schneider, Zschiegner,
  Wang, and Wang]{kollovieh2023predict}
Marcel Kollovieh, Abdul~Fatir Ansari, Michael Bohlke-Schneider, Jasper
  Zschiegner, Hao Wang, and Yuyang Wang.
\newblock Predict, refine, synthesize: Self-guiding diffusion models for
  probabilistic time series forecasting.
\newblock \emph{arXiv preprint arXiv:2307.11494}, 2023.

\bibitem[Kong et~al.(2020)Kong, Ping, Huang, Zhao, and
  Catanzaro]{kong2020diffwave}
Zhifeng Kong, Wei Ping, Jiaji Huang, Kexin Zhao, and Bryan Catanzaro.
\newblock Diffwave: A versatile diffusion model for audio synthesis.
\newblock In \emph{International Conference on Learning Representations}, 2020.

\bibitem[Krizhevsky et~al.(2009)Krizhevsky, Hinton,
  et~al.]{krizhevsky2009learning}
Alex Krizhevsky, Geoffrey Hinton, et~al.
\newblock Learning multiple layers of features from tiny images.
\newblock 2009.

\bibitem[Laidlaw et~al.(2020)Laidlaw, Singla, and Feizi]{laidlaw2020perceptual}
Cassidy Laidlaw, Sahil Singla, and Soheil Feizi.
\newblock Perceptual adversarial robustness: Defense against unseen threat
  models.
\newblock \emph{arXiv preprint arXiv:2006.12655}, 2020.

\bibitem[Lienen et~al.(2023)Lienen, Hansen-Palmus, L{\"u}dke, and
  G{\"u}nnemann]{lienen2023generative}
Marten Lienen, Jan Hansen-Palmus, David L{\"u}dke, and Stephan G{\"u}nnemann.
\newblock Generative diffusion for 3d turbulent flows.
\newblock \emph{arXiv preprint arXiv:2306.01776}, 2023.

\bibitem[Lin et~al.(2019)Lin, Song, He, Wang, and Hopcroft]{lin2019nesterov}
Jiadong Lin, Chuanbiao Song, Kun He, Liwei Wang, and John~E Hopcroft.
\newblock Nesterov accelerated gradient and scale invariance for adversarial
  attacks.
\newblock \emph{arXiv preprint arXiv:1908.06281}, 2019.

\bibitem[Madry et~al.(2017)Madry, Makelov, Schmidt, Tsipras, and
  Vladu]{madry2017towards}
Aleksander Madry, Aleksandar Makelov, Ludwig Schmidt, Dimitris Tsipras, and
  Adrian Vladu.
\newblock Towards deep learning models resistant to adversarial attacks.
\newblock \emph{arXiv preprint arXiv:1706.06083}, 2017.

\bibitem[Meng et~al.(2021)Meng, He, Song, Song, Wu, Zhu, and
  Ermon]{meng2021sdedit}
Chenlin Meng, Yutong He, Yang Song, Jiaming Song, Jiajun Wu, Jun-Yan Zhu, and
  Stefano Ermon.
\newblock Sdedit: Guided image synthesis and editing with stochastic
  differential equations.
\newblock In \emph{International Conference on Learning Representations}, 2021.

\bibitem[Narodytska \& Kasiviswanathan(2016)Narodytska and
  Kasiviswanathan]{narodytska2016simple}
Nina Narodytska and Shiva~Prasad Kasiviswanathan.
\newblock Simple black-box adversarial perturbations for deep networks.
\newblock \emph{arXiv preprint arXiv:1612.06299}, 2016.

\bibitem[Nichol et~al.(2021)Nichol, Dhariwal, Ramesh, Shyam, Mishkin, McGrew,
  Sutskever, and Chen]{nichol2021glide}
Alex Nichol, Prafulla Dhariwal, Aditya Ramesh, Pranav Shyam, Pamela Mishkin,
  Bob McGrew, Ilya Sutskever, and Mark Chen.
\newblock Glide: Towards photorealistic image generation and editing with
  text-guided diffusion models.
\newblock \emph{arXiv preprint arXiv:2112.10741}, 2021.

\bibitem[Nie et~al.(2022)Nie, Guo, Huang, Xiao, Vahdat, and
  Anandkumar]{nie2022DiffPure}
Weili Nie, Brandon Guo, Yujia Huang, Chaowei Xiao, Arash Vahdat, and Anima
  Anandkumar.
\newblock Diffusion models for adversarial purification.
\newblock In \emph{International Conference on Machine Learning (ICML)}, 2022.

\bibitem[Peebles \& Xie(2022)Peebles and Xie]{Peebles2022DiT}
William Peebles and Saining Xie.
\newblock Scalable diffusion models with transformers.
\newblock \emph{arXiv preprint arXiv:2212.09748}, 2022.

\bibitem[Peng et~al.(2023)Peng, Xu, Cornelius, Hull, Li, Duggal, Phute, Martin,
  and Chau]{peng2023robust}
ShengYun Peng, Weilin Xu, Cory Cornelius, Matthew Hull, Kevin Li, Rahul Duggal,
  Mansi Phute, Jason Martin, and Duen~Horng Chau.
\newblock Robust principles: Architectural design principles for adversarially
  robust cnns.
\newblock \emph{arXiv preprint arXiv:2308.16258}, 2023.

\bibitem[Salman et~al.(2020)Salman, Ilyas, Engstrom, Kapoor, and
  Madry]{salman2020adversarially}
Hadi Salman, Andrew Ilyas, Logan Engstrom, Ashish Kapoor, and Aleksander Madry.
\newblock Do adversarially robust imagenet models transfer better?
\newblock \emph{Advances in Neural Information Processing Systems},
  33:\penalty0 3533--3545, 2020.

\bibitem[Samangouei et~al.(2018)Samangouei, Kabkab, and
  Chellappa]{samangouei2018defense}
Pouya Samangouei, Maya Kabkab, and Rama Chellappa.
\newblock Defense-gan: Protecting classifiers against adversarial attacks using
  generative models.
\newblock \emph{arXiv preprint arXiv:1805.06605}, 2018.

\bibitem[Smith \& Topin(2019)Smith and Topin]{smith2019super}
Leslie~N Smith and Nicholay Topin.
\newblock Super-convergence: Very fast training of neural networks using large
  learning rates.
\newblock In \emph{Artificial intelligence and machine learning for
  multi-domain operations applications}, volume 11006, pp.\  369--386. SPIE,
  2019.

\bibitem[Sohl-Dickstein et~al.(2015)Sohl-Dickstein, Weiss, Maheswaranathan, and
  Ganguli]{sohl2015deep}
Jascha Sohl-Dickstein, Eric Weiss, Niru Maheswaranathan, and Surya Ganguli.
\newblock Deep unsupervised learning using nonequilibrium thermodynamics.
\newblock In \emph{International conference on machine learning}, pp.\
  2256--2265. PMLR, 2015.

\bibitem[Song et~al.(2017)Song, Kim, Nowozin, Ermon, and
  Kushman]{song2017pixeldefend}
Yang Song, Taesup Kim, Sebastian Nowozin, Stefano Ermon, and Nate Kushman.
\newblock Pixeldefend: Leveraging generative models to understand and defend
  against adversarial examples.
\newblock \emph{arXiv preprint arXiv:1710.10766}, 2017.

\bibitem[Song et~al.(2018)Song, Shu, Kushman, and Ermon]{song2018unrestricted}
Yang Song, Rui Shu, Nate Kushman, and Stefano Ermon.
\newblock Constructing unrestricted adversarial examples with generative
  models.
\newblock In \emph{Advances in Neural Information Processing Systems},
  volume~31, 2018.

\bibitem[Song et~al.(2020)Song, Sohl-Dickstein, Kingma, Kumar, Ermon, and
  Poole]{song2020score}
Yang Song, Jascha Sohl-Dickstein, Diederik~P Kingma, Abhishek Kumar, Stefano
  Ermon, and Ben Poole.
\newblock Score-based generative modeling through stochastic differential
  equations.
\newblock \emph{arXiv preprint arXiv:2011.13456}, 2020.

\bibitem[Su et~al.(2019)Su, Vargas, and Sakurai]{su2019one}
Jiawei Su, Danilo~Vasconcellos Vargas, and Kouichi Sakurai.
\newblock One pixel attack for fooling deep neural networks.
\newblock \emph{IEEE Transactions on Evolutionary Computation}, 23\penalty0
  (5):\penalty0 828--841, 2019.

\bibitem[Tram{\`e}r et~al.(2020)Tram{\`e}r, Behrmann, Carlini, Papernot, and
  Jacobsen]{tramer2020fundamental}
Florian Tram{\`e}r, Jens Behrmann, Nicholas Carlini, Nicolas Papernot, and
  J{\"o}rn-Henrik Jacobsen.
\newblock Fundamental tradeoffs between invariance and sensitivity to
  adversarial perturbations.
\newblock In \emph{International Conference on Machine Learning}, pp.\
  9561--9571. PMLR, 2020.

\bibitem[Wang(2021)]{wang2021adversarial}
Jiakai Wang.
\newblock Adversarial examples in physical world.
\newblock In \emph{IJCAI}, pp.\  4925--4926, 2021.

\bibitem[Wang et~al.(2023)Wang, Pang, Du, Lin, Liu, and Yan]{wang2023better}
Zekai Wang, Tianyu Pang, Chao Du, Min Lin, Weiwei Liu, and Shuicheng Yan.
\newblock Better diffusion models further improve adversarial training.
\newblock \emph{arXiv preprint arXiv:2302.04638}, 2023.

\bibitem[Xie et~al.(2019)Xie, Zhang, Zhou, Bai, Wang, Ren, and
  Yuille]{xie2019improving}
Cihang Xie, Zhishuai Zhang, Yuyin Zhou, Song Bai, Jianyu Wang, Zhou Ren, and
  Alan~L Yuille.
\newblock Improving transferability of adversarial examples with input
  diversity.
\newblock In \emph{Proceedings of the IEEE/CVF conference on computer vision
  and pattern recognition}, pp.\  2730--2739, 2019.

\bibitem[Xue et~al.(2023)Xue, Araujo, Hu, and Chen]{xue2023diffusion}
Haotian Xue, Alexandre Araujo, Bin Hu, and Yongxin Chen.
\newblock Diffusion-based adversarial sample generation for improved
  stealthiness and controllability.
\newblock \emph{arXiv preprint arXiv:2305.16494}, 2023.

\bibitem[Yoon et~al.(2021)Yoon, Hwang, and Lee]{yoon2021adversarial}
Jongmin Yoon, Sung~Ju Hwang, and Juho Lee.
\newblock Adversarial purification with score-based generative models.
\newblock In \emph{International Conference on Machine Learning}, pp.\
  12062--12072. PMLR, 2021.

\bibitem[Zagoruyko \& Komodakis(2016)Zagoruyko and
  Komodakis]{zagoruyko2016wide}
Sergey Zagoruyko and Nikos Komodakis.
\newblock Wide residual networks.
\newblock \emph{arXiv preprint arXiv:1605.07146}, 2016.

\bibitem[Zhang et~al.(2019)Zhang, Yu, Jiao, Xing, El~Ghaoui, and
  Jordan]{zhang2019theoretically}
Hongyang Zhang, Yaodong Yu, Jiantao Jiao, Eric Xing, Laurent El~Ghaoui, and
  Michael Jordan.
\newblock Theoretically principled trade-off between robustness and accuracy.
\newblock In \emph{International conference on machine learning}, pp.\
  7472--7482. PMLR, 2019.

\bibitem[Zhang et~al.(2018)Zhang, Isola, Efros, Shechtman, and
  Wang]{zhang2018unreasonable}
Richard Zhang, Phillip Isola, Alexei~A Efros, Eli Shechtman, and Oliver Wang.
\newblock The unreasonable effectiveness of deep features as a perceptual
  metric.
\newblock In \emph{Proceedings of the IEEE conference on computer vision and
  pattern recognition}, pp.\  586--595, 2018.

\end{thebibliography}
\bibliographystyle{tmlr}

\newpage
\appendix

\section{Experimental Setup and Hyperparameters}
\subsection{Reproducibility}
Our models are implemented using PyTorch with the pre-trained EDM models by \citet{karras2022elucidating} and \citet{wang2023better}, and the guidance scores are computed using automatic differentiation. In Tab.~\ref{tab:classifier_params} and Tab.~\ref{tab:generative_params}, we give an overview of the hyperparameters of \oursacro{}. For the methods DiffAttack, DiffPure, CAA, PPGD, and LPA, we use the corresponding authors' official implementations with the suggested hyperparameters. For the remaining attacks, we use Adversarial-Attacks-PyTorch with its default parameters \citep{kim2020torchattacks}.
\subsection{Hyperparameters}
To train the WRN-28-10 classifiers, we use the parameters shown in Tab.~\ref{tab:classifier_params}. In Tab.~\ref{tab:generative_params}, we show the scale parameters used to evaluate the attacks and purification of \oursacro{}, i.e., the results shown in Tab.~\ref{tab:generation}, Tab.~\ref{tab:purification} and Tab.~\ref{tab:generation_additional}. The attacks on robust models do not sequentially increase the scale $s_\mathbf{y}$ but use fixed scales of $s_\mathbf{x}=48$ and $s_\mathbf{y}=80$. For the common corruptions we use a scale of $s_\mathbf{x}=40$ on the robust models. Finally, for the EDM sampler we use the default sampling scheduler and parameters by \citet{karras2022elucidating}.
\begin{table}[h]
    \small
    \centering
    \begin{tabular}{lc}
        Hyperparameter & Value \\
        \midrule
        Number of epochs & 400\\
          Optimizer & SGD\\
          Nesterov momentum & 0.9 \\
          Weight decay & $5\times 10^{-4}$ \\
          Exponential moving average & 0.995\\
          Learning rate scheduler & Cyclic with cosine annealing \\
          Initial learning rate & 0.2 \\
    \end{tabular}
    \caption{Hyperparameters used to train the WRN-28-10 classifiers.}
    \label{tab:classifier_params}
\end{table}

\begin{table}[h]
    \scriptsize
    \centering
    \begin{tabular}{lc}
        Hyperparameter & Value \\
        \midrule
        \multicolumn{1}{l}{\cellcolor{Gray}\textbf{CIFAR10}} & \multicolumn{1}{c}{\cellcolor{Gray}}\\
        \hspace{2mm}$s_\mathbf{y}$ (GAT) & 32\\
        \hspace{2mm}$s_\mathbf{x}$ (GAT) & 48\\
        \hspace{2mm}$s_\mathbf{y}$ (GAT-LPIPS) & 32\\
        \hspace{2mm}$s_\mathbf{x}$ (GAT-LPIPS) & 48\\
        \hspace{2mm}$s_\mathrm{LPIPS}$ (GAT-LPIPS) & 48\\
        \hspace{2mm}$s_\mathbf{x}$ (GAP) & 10 \\
        \hspace{2mm}increments (GAT) & 20 \\
        \hspace{2mm}steps (GAP) & 72 \\
        \hspace{2mm}steps (GAT) & 512 \\
        \multicolumn{1}{l}{\cellcolor{Gray}\textbf{CIFAR100}} & \multicolumn{1}{c}{\cellcolor{Gray}}\\
        \hspace{2mm}$s_\mathbf{y}$ (GAT) & 32\\
        \hspace{2mm}$s_\mathbf{x}$ (GAT) & 48\\
        \hspace{2mm}$s_\mathbf{y}$ (GAT-LPIPS) & 32\\
        \hspace{2mm}$s_\mathbf{x}$ (GAT-LPIPS) & 48\\
        \hspace{2mm}$s_\mathrm{LPIPS}$ (GAT-LPIPS) & 48\\
        \hspace{2mm}increments & 20 \\
        \hspace{2mm}steps & 512 \\
        \multicolumn{1}{l}{\cellcolor{Gray}\textbf{TinyImagenet}} & \multicolumn{1}{c}{\cellcolor{Gray}}\\
        \hspace{2mm}$s_\mathbf{y}$ (GAT) & 64\\
        \hspace{2mm}$s_\mathbf{x}$ (GAT) & 16\\
        \hspace{2mm}increments & 20 \\
        \hspace{2mm}steps & 512 \\
      {\cellcolor{Gray}\textbf{Imagenet-Compatible}} & \multicolumn{1}{c}{\cellcolor{Gray}}\\
        \hspace{2mm}$s_\mathbf{y}$ (GAT) & 8\\
        \hspace{2mm}$s_\mathbf{x}$ (GAT) & 0.5\\
        \hspace{2mm}increments & 4 \\
        \hspace{2mm}steps & 1000 \\
    \end{tabular}
    \caption{Hyperparameters used to evaluate \oursacro{}.}
    \label{tab:generative_params}
\end{table}
\begin{table}[h]
    \small
    \centering
    \begin{tabular}{lc}
        Hyperparameter & Value \\
        \midrule
          $\sigma_{\min}$ & $0.002$\\
          $\sigma_{\max}$ & $80$ \\
          $\rho$ & $7$ \\
          $S_\mathrm{churn}$ & 0/4 \\
          $S_\mathrm{noise}$ & 1 \\
    \end{tabular}
    \caption{Hyperparameters used for sampling using Alg.~\ref{alg:pseudo}.}
    \label{tab:sampling_params}
\end{table}
\subsection{Pseudocode}
We present the pseudocode of \oursacro{} in Alg.~\ref{alg:pseudo}, implementing the sampler proposed by \citet{karras2022elucidating}. Here, $s$ denotes the scale parameter for the task, while $t_i$ and $\gamma_i$ are scheduler parameters retained from the original configuration (see Tab.~\ref{tab:sampling_params}). More specifically, $\gamma_i=\min(S_\mathrm{churn},\sqrt{2}-1)$ and
\begin{equation}
    t_{i < N} = \left( \sigma_{\text{max}}^{\frac{1}{\rho}} + \frac{i}{N - 1} \left( \sigma_{\text{min}}^{\frac{1}{\rho}} - \sigma_{\text{max}}^{\frac{1}{\rho}} \right) \right)^{\rho}, \quad t_N = 0.
\end{equation}
We compute the different guidance scores using \eqref{eq:synthesis}, \eqref{eq:transform}, and \eqref{eq:purification}.
\begin{algorithm}
\caption{\oursacro{} with the sampler of \citet{karras2022elucidating}.}
\begin{algorithmic}[1]
\Procedure{\oursacro{}}{$s_\theta(\mathbf{x}; \sigma), t_{i \in \{0, \dots, N\}}, \gamma_{i \in \{0, \dots, N-1\}}, s, c$}
    \State sample $\mathbf{x}_0 \sim \mathcal{N}(0, t_0^2 \mathbf{I})$
    \For{$i \in \{0, \dots, N-1\}$}
        \State sample $\epsilon_i \sim \mathcal{N}(0, S_\mathrm{noise}^2\mathbf{I})$
        \State $\hat{t}_i \gets t_i + \gamma_i t_i$
        \State $\hat{\mathbf{x}}_i \gets \mathbf{x}_i + \sqrt{\hat{t}_i^2 - t_i^2} \, \epsilon_i$
        \State $d_i \gets  \hat{t}_i \cdot \big(s_\theta(\mathbf{x}_i, \hat{t}_i) + s\cdot \nabla_{\mathbf{x}_i} \log p_{\hat{t}_i}(c\mid\mathbf{x}_i)\big)$
        \State $\mathbf{x}_{i+1} \gets \hat{\mathbf{x}}_i + (t_{i+1} - \hat{t}_i) d_i$
        \If{$t_{i+1} \neq 0$}
            \State $d_i' \gets t_{i+1} \cdot \big(s_\theta(\mathbf{x}_{i+1}, t_{i+1}) + s\cdot \nabla_{\mathbf{x}_{i+1}} \log p_{t_{i+1}}(c\mid\mathbf{x}_{i+1})\big)$
            \State $\mathbf{x}_{i+1} \gets \hat{\mathbf{x}}_i + (t_{i+1} - \hat{t}_i) \left(\frac{1}{2} d_i + \frac{1}{2} d_i'\right)$
        \EndIf
    \EndFor
    \State \textbf{return} $\mathbf{x}_N$
\EndProcedure
\label{alg:pseudo}
\end{algorithmic}
\end{algorithm}
\clearpage
\section{Additional Results}
\subsection{Qualitative Comparison of Baselines}
\begin{figure*}[h]
    \centering
 \begin{minipage}{0.44\linewidth}
    \includegraphics[width=\cifarwidth px]{figures/cifar_examples/clean1.png}
    \includegraphics[width=\cifarwidth px]{figures/cifar_examples/None_48_48_1_label_8_pred_1_0.40276917815208435_0.04399070888757706.png}
    \includegraphics[width=\cifarwidth px]{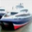}
    \includegraphics[width=\cifarwidth px]{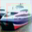}
    \includegraphics[width=\cifarwidth px]{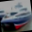}
    \includegraphics[width=\cifarwidth px]{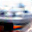} \\
    \includegraphics[width=\cifarwidth px]{figures/cifar_examples/clean13.png}
    \includegraphics[width=\cifarwidth px]{figures/cifar_examples/None_48_48_13_label_7_pred_4_0.6886034607887268_0.07536858320236206.png}
        \includegraphics[width=\cifarwidth px]{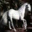}
    \includegraphics[width=\cifarwidth px]{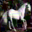}
    \includegraphics[width=\cifarwidth px]{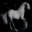}
    \includegraphics[width=\cifarwidth px]{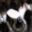} \\
    \includegraphics[width=\cifarwidth px]{figures/cifar_examples/clean24.png}
    \includegraphics[width=\cifarwidth px]{figures/cifar_examples/None_48_48_24_label_5_pred_3_0.4763997793197632_0.0443190336227417.png}
        \includegraphics[width=\cifarwidth px]{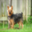}
    \includegraphics[width=\cifarwidth px]{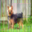}
    \includegraphics[width=\cifarwidth px]{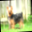}
    \includegraphics[width=\cifarwidth px]{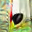} \\
        \begin{minipage}{0.152\linewidth}
        \scriptsize
        \centering
        \smash{Original}
        \end{minipage} 
        \begin{minipage}{0.152\linewidth}
        \scriptsize
        \centering
        \smash{\oursacro{}}       
        \end{minipage} 
        \begin{minipage}{0.152\linewidth}
        \scriptsize
        \centering
        \smash{FAB}       
        \end{minipage} 
        \begin{minipage}{0.152\linewidth}
        \scriptsize
        \centering
        \smash{Square}       
        \end{minipage} 
        \begin{minipage}{0.152\linewidth}
        \scriptsize
        \centering
        \smash{CAA}       
        \end{minipage} 
        \begin{minipage}{0.152\linewidth}
        \scriptsize
        \centering
        \smash{DiffAttack}       
        \end{minipage} \\
        \vspace{-0.5em}
\end{minipage}
\vspace{-1em}
\caption{Examples from the CIFAR10 dataset. The figure presents selected baseline images corresponding to the examples in Fig.~\ref{fig:cifar_transformation}. For \oursacro{}-GAT, we used $s_\mathbf{x}=48$. As baselines, we included FAB ($\ell_2=0.5$) and Square ($\ell_\infty=8/255$) to represent restricted attacks, as they achieve the lowest and highest LPIPS scores, respectively. Additionally, we show the two unrestricted baselines, CAA and DiffAttack.}
\vspace{-1.5em}
\label{fig:cifar_baselines}
\end{figure*}

In Fig.~\ref{fig:cifar_baselines}, we visualize adversarial attacks of selected baselines for the images in Fig.~\ref{fig:cifar_main}.

\subsection{Qualitative Effect of the Scale Parameters}
\begin{figure*}[h!]
    \centering
    \subfigure[$s_\mathbf{x}=0$]{
\includegraphics[width=0.15\textwidth]{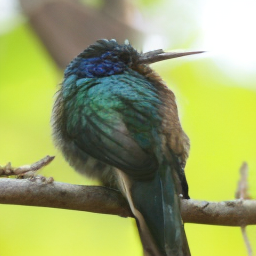}} 
    \subfigure[$s_\mathbf{x}=0.125$]{
\includegraphics[width=0.15\textwidth]{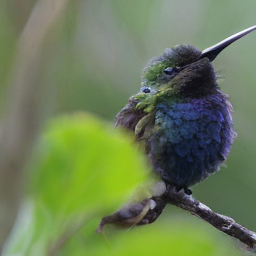}}
    \subfigure[$s_\mathbf{x}=0.25$]{
\includegraphics[width=0.15\textwidth]{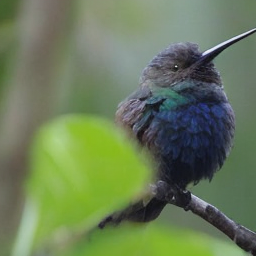}}
\subfigure[$s_\mathbf{x}=0.5$]{
\includegraphics[width=0.15\textwidth]{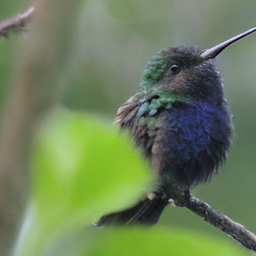}}
\subfigure[$s_\mathbf{x}=1.0$]{
\includegraphics[width=0.15\textwidth]{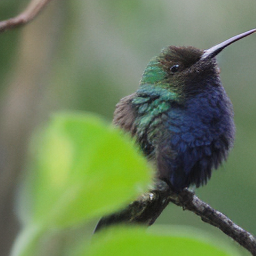}}
\subfigure[Original]{\includegraphics[width=0.15\textwidth]{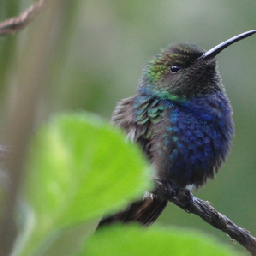}} 
    \vspace{-0.5em}
    \caption{Effect of the scale parameter $s_\mathbf{x}$. The images display adversarial images generated by \oursacro{}-GAT across different scales $s_\mathbf{x}$ on a robust WRN-50-2~\citep{salman2020adversarially} with $s_\mathbf{y}=8$. For $s_\mathbf{x}=0$, the setup equals the GAS setup and synthesizes an image unrelated to the input. As the scale increases, the image gets closer to the original.}
    \label{fig:x_scales}
\end{figure*}

\begin{figure*}[h!]
    \centering
    \subfigure[Original]{\includegraphics[width=0.15\textwidth]{figures/imagenet_compatible/img_78_original.png}}
    \subfigure[$s_\mathbf{y}=0$]{
\includegraphics[width=0.15\textwidth]{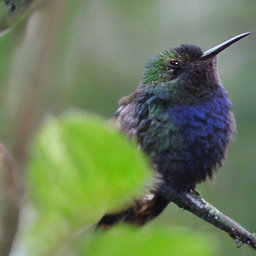}} 
    \subfigure[$s_\mathbf{y}=8$]{
\includegraphics[width=0.15\textwidth]{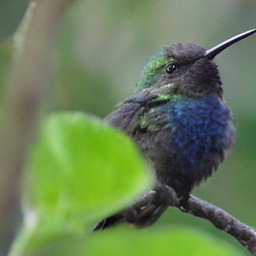}}
    \subfigure[$s_\mathbf{y}=16$]{
\includegraphics[width=0.15\textwidth]{figures/imagenet_compatible/img_78_94_95_0.25_16_2000_32.128204345703125_827_mvnet_icml.png}}
\subfigure[$s_\mathbf{y}=32$]{
\includegraphics[width=0.15\textwidth]{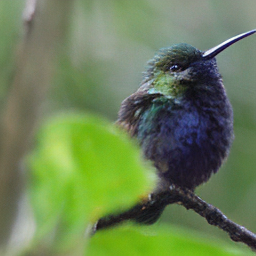}}
\subfigure[$s_\mathbf{y}=48$]{
\includegraphics[width=0.15\textwidth]{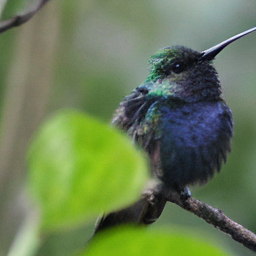}}
    \vspace{-0.5em}
    \caption{Effect of the scale parameter $s_\mathbf{y}$. The images display adversarial images generated by \oursacro{}-GAT across different scales $s_\mathbf{y}$ on a robust WRN-50-2~\citep{salman2020adversarially} with $s_\mathbf{x}=0.25$. For $s_\mathbf{y}=0$, the setup equals the GAP setup and synthesizes an image without adversarial perturbations. As the scale increases, the adversarial content strengthens, causing the images to diverge further from the original.}
    \label{fig:y_scales}
\end{figure*}

To provide a more intuitive understanding of \oursacro{}, we show the visual effect of the scale parameters $s_\mathbf{x}$ and $s_\mathbf{y}$ in Fig.~\ref{fig:x_scales} and \ref{fig:y_scales}. These visualizations illustrate the effects of the scale parameters $s_\mathbf{x}$ and $s_\mathbf{y}$. When both scale parameters are set to zero, the model behaves as a standard diffusion model. Increasing $s_\mathbf{x}$ guides the diffusion process toward a specific image, which is used in the GAP setup. Increasing $s_\mathbf{y}$ introduces adversarial perturbations, allowing the synthesis of adversarial images. When both parameters are greater than zero, the GAT model transforms existing images into adversarial examples. 

\subsection{Additional classifiers for adversarial attacks using GAT}\label{sec:additional_classifiers}
To verify the efficacy of \oursacro{} and demonstrate its applicability across various architectures, we evaluate the accuracy of GAT on four more pretrained classifiers via \texttt{PyTorch Hub}\footnote{\href{https://github.com/chenyaofo/pytorch-cifar-models}{https://github.com/chenyaofo/pytorch-cifar-models}} for the datasets CIFAR10 and CIFAR100 using the same hyperparameters, i.e., scale parameters, as for the WRN-28-10 classifier. We show the adversarial accuracy in Tab.~\ref{tab:generation_additional}, including selected baselines. As we can observe, \oursacro{} successfully generates adversarial attacks on various classifiers, reaching accuracies close to 0\%. This demonstrates the flexibility of \oursacro{} and applicability to arbitrary pre-trained classifiers.
\begin{table*}[!ht]
\caption{Adversarial accuracy of \oursacro{} for various classifiers on the datasets CIFAR10 and CIFAR100.}
\vspace{0.5em}
\label{tab:generation_additional}
    \scriptsize
    \centering
    \resizebox{\linewidth}{!}{
    \begin{tabular}{l c c c c c c c c} 
    \arrayrulecolor{black}
    & \multicolumn{2}{c}{ResNet-20} & \multicolumn{2}{c}{ResNet-56} & \multicolumn{2}{c}{VGG-19} & \multicolumn{2}{c}{RepVGG-A2} \\
Dataset    & CIFAR10 & CIFAR100 & CIFAR10 & CIFAR100 & CIFAR10 & CIFAR100 & CIFAR10 & CIFAR100 \\
    \midrule
   \multicolumn{1}{l}{\cellcolor{Gray}\textbf{$\mathbf{\ell_\infty}$ restricted}} & \multicolumn{8}{c}{\cellcolor{Gray}}\\
    \hspace{3mm}FGSM~\citep{goodfellow2014explaining} & 14.95 & 4.97 & 34.33 & 8.09 & 29.96 & 20.87 & 51.37 & 10.20\\
    \hspace{3mm}DI-FGSM~\citep{xie2019improving} & \textbf{0.00} & 0.01 & 0.12  & \textbf{0.00} & 1.06 & 2.17 & 1.55 & 2.17\\
    \hspace{3mm}SI-NI-FGSM~\citep{lin2019nesterov} & 0.51 & 0.14 & 2.12 & 0.55 & 11.89 & 4.74 & 4.27 & 4.74\\
    \hspace{3mm}APGD~\citep{croce2020reliable} & \textbf{0.00} & 0.01 & 0.01 & 0.01 & 0.14 & 0.77 & 0.06 & 0.77\\
    \hspace{3mm}APGDT~\citep{croce2020reliable} & \textbf{0.00} & \textbf{0.00} & \textbf{0.00} & \textbf{0.00} & \textbf{0.00} & \textbf{0.00} & \textbf{0.00} & \textbf{0.00}\\
    \hspace{3mm}Square~\citep{andriushchenko2020square} & \textbf{0.00} & \textbf{0.00} & \textbf{0.00} & \textbf{0.00} & 0.51 & 0.76 & 0.42 & 0.11\\
    \hspace{3mm}FAB~\citep{croce2020minimally} & 0.29 & 0.36 & 0.31 & 0.35 & 4.79 & 2.67 & 1.58 & 0.18\\
    \multicolumn{1}{l}{\cellcolor{Gray}\textbf{$\mathbf{\ell_2}$ restricted}} & \multicolumn{8}{c}{\cellcolor{Gray}}\\ 
    \hspace{3mm}APGD~\citep{croce2020reliable} & \textbf{0.00} & 0.01 & 0.01 & 0.01 & 0.14 & 0.77 & 0.06 & 0.02\\
    \hspace{3mm}APGDT~\citep{croce2020reliable} & \textbf{0.00} & \textbf{0.00} & \textbf{0.00} & \textbf{0.00} & \textbf{0.00} & \textbf{0.00} & \textbf{0.00} & \textbf{0.00}\\
    \hspace{3mm}Square~\citep{andriushchenko2020square} & \textbf{0.00} & \textbf{0.00} & \textbf{0.00} & \textbf{0.00} & 0.51 & 0.76 & 0.42 & 0.11 \\
    \hspace{3mm}FAB~\citep{croce2020minimally} & 0.25 & 0.40 & 0.30 & 0.31 & 4.79 & 2.67 & 1.53 & 0.14 \\

    \multicolumn{1}{l}{\cellcolor{Gray}\textbf{$\mathbf{\ell_0}$ restricted}} & \multicolumn{8}{c}{\cellcolor{Gray}}\\
    \hspace{3mm}OnePixel~\citep{su2019one} & 76.39 & 42.28 & 81.00 & 44.60 & 74.49 & 45.47 & 82.79 & 56.17 \\
    \multicolumn{1}{l}{\cellcolor{Gray}\textbf{Unrestricted}} & \multicolumn{8}{c}{\cellcolor{Gray}}\\
     \hspace{3mm}CAA~\citep{hsiung2022caa} & 25.10 & 4.05 & 36.75 & 5.16 & 33.75 & 10.20 & 43.16 & 7.43 \\
   \hspace{3mm}PPGD~\citep{laidlaw2020perceptual} & 50.93 & 29.76 & 43.88 & 32.52 & 10.95 & 23.33 & 33.14 & 33.53 \\
    \hspace{3mm}LPA~\citep{laidlaw2020perceptual} & \textbf{0.00} & \textbf{0.00} & \textbf{0.00} & \textbf{0.00} & \underline{0.02} & 0.30 & \underline{0.01} & \textbf{0.00} \\
    \hspace{3mm}\oursacro{} (Ours) & \textbf{0.00} & \textbf{0.00} & \textbf{0.00} & \textbf{0.00} & \underline{0.02} & \underline{0.16} & \textbf{0.00} & \textbf{0.00} \\
    \bottomrule
\end{tabular}}
\end{table*}

\begin{figure*}[h]
 \begin{minipage}{\linewidth}
     \centering
\includegraphics[width=\imagenetwidth px]{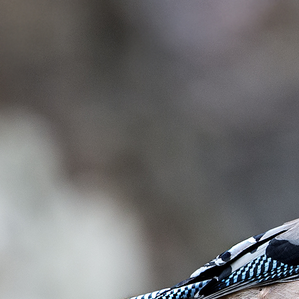}
\includegraphics[width=\imagenetwidth px]{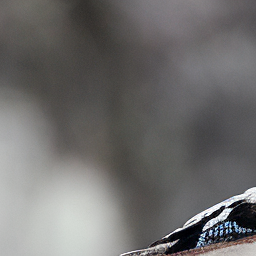}
\includegraphics[width=\imagenetwidth px]{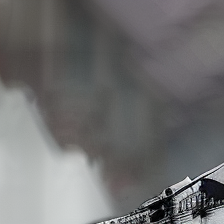}
\includegraphics[width=\imagenetwidth px]{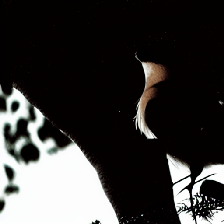}
\includegraphics[width=\imagenetwidth px]{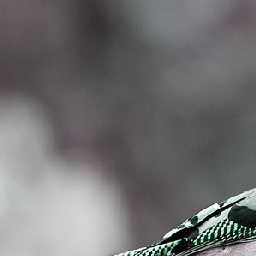}
\includegraphics[width=\imagenetwidth px]{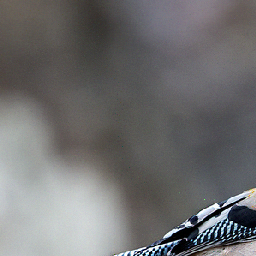}
\\
\includegraphics[width=\imagenetwidth px]{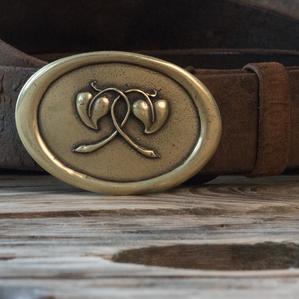}
\includegraphics[width=\imagenetwidth px]{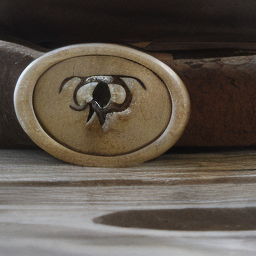}
\includegraphics[width=\imagenetwidth px]{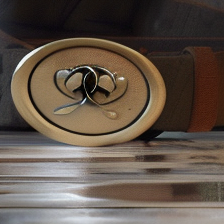}
\includegraphics[width=\imagenetwidth px]{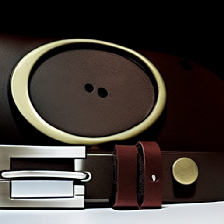}
\includegraphics[width=\imagenetwidth px]{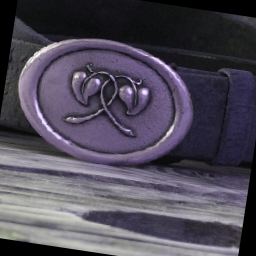}
\includegraphics[width=\imagenetwidth px]{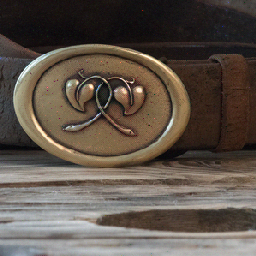}
\\
\includegraphics[width=\imagenetwidth px]{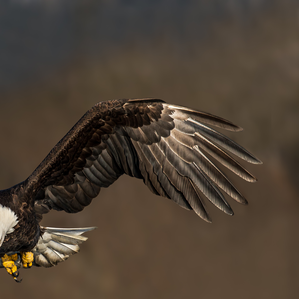}
\includegraphics[width=\imagenetwidth px]{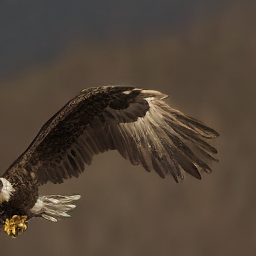}
\includegraphics[width=\imagenetwidth px]{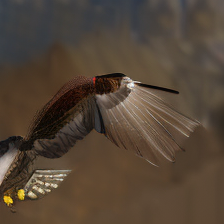}
\includegraphics[width=\imagenetwidth px]{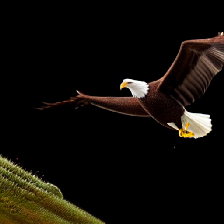}
\includegraphics[width=\imagenetwidth px]{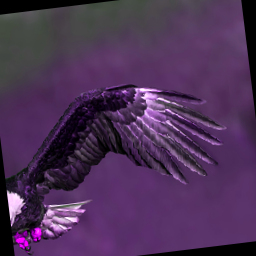}
\includegraphics[width=\imagenetwidth px]{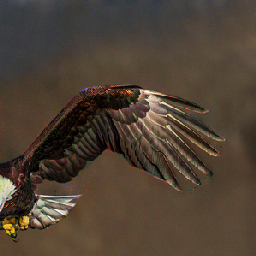}
\\
\includegraphics[width=\imagenetwidth px]{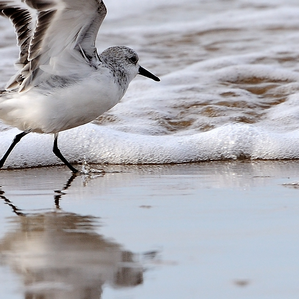}
\includegraphics[width=\imagenetwidth px]{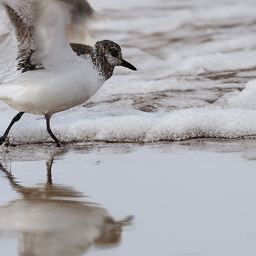}
\includegraphics[width=\imagenetwidth px]{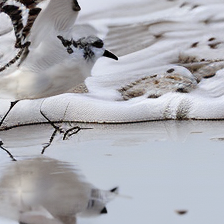}
\includegraphics[width=\imagenetwidth px]{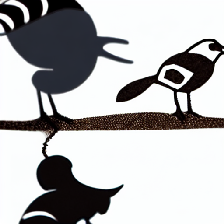}
\includegraphics[width=\imagenetwidth px]{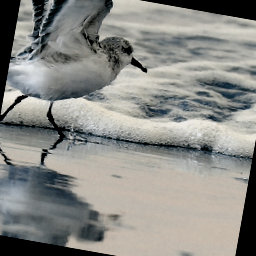}
\includegraphics[width=\imagenetwidth px]{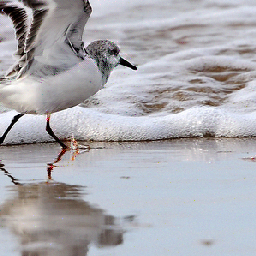}
\\
\includegraphics[width=\imagenetwidth px]{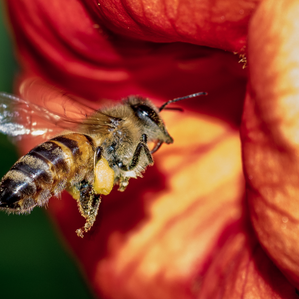}
\includegraphics[width=\imagenetwidth px]{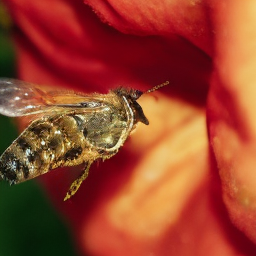}
\includegraphics[width=\imagenetwidth px]{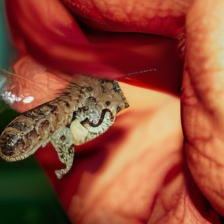}
\includegraphics[width=\imagenetwidth px]{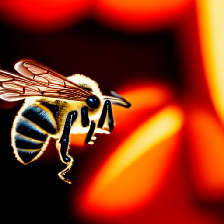}
\includegraphics[width=\imagenetwidth px]{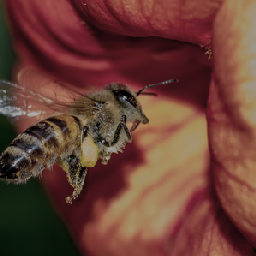}
\includegraphics[width=\imagenetwidth px]{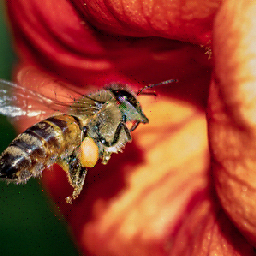}
\\
\includegraphics[width=\imagenetwidth px]{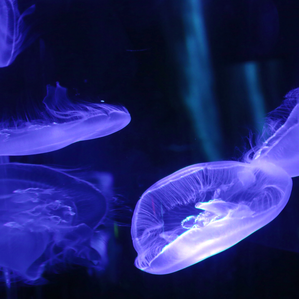}
\includegraphics[width=\imagenetwidth px]{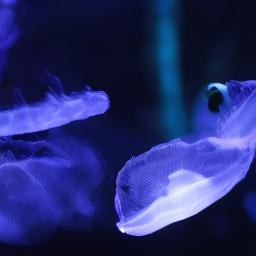}
\includegraphics[width=\imagenetwidth px]{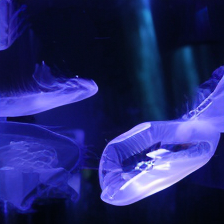}
\includegraphics[width=\imagenetwidth px]{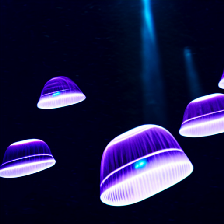}
\includegraphics[width=\imagenetwidth px]{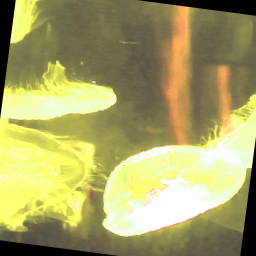}
\includegraphics[width=\imagenetwidth px]{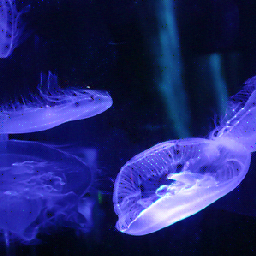}
\\
\includegraphics[width=\imagenetwidth px]{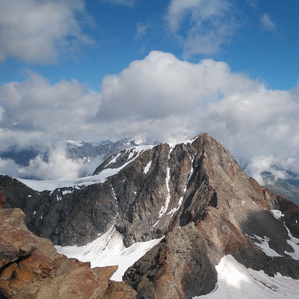}
\includegraphics[width=\imagenetwidth px]{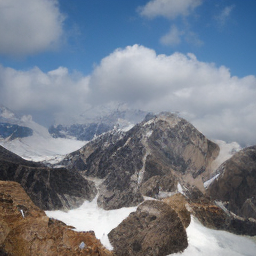}
\includegraphics[width=\imagenetwidth px]{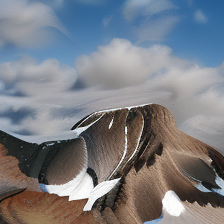}
\includegraphics[width=\imagenetwidth px]{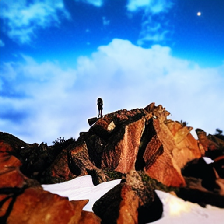}
\includegraphics[width=\imagenetwidth px]{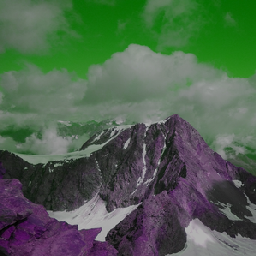}
\includegraphics[width=\imagenetwidth px]{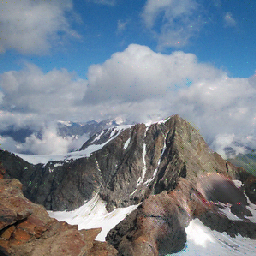}
\end{minipage}
    \centering
\subfigure[Original.]{\hspace{.130\textwidth}} 
\subfigure[\oursacro{}.]{\hspace{.130\textwidth}} 
\subfigure[DiffAttack.]{\hspace{.130\textwidth}} 
\subfigure[ACA.]{\hspace{.130\textwidth}} 
\subfigure[CAA.]{\hspace{.130\textwidth}} 
\subfigure[LPA.]{\hspace{.130\textwidth}} 
\vspace{-0.5em}
\caption{Adversarial examples on the ImageNet-Compatible dataset of various classes for different unrestricted attacks.
}
\label{fig:imagenet_compatible}
\end{figure*}

Additionally, we evaluate \oursacro{} on the high-resolution ImageNet-Compatible\footnote{\url{https://github.com/cleverhans-lab/cleverhans/tree/master/cleverhans_v3.1.0/ examples/nips17_adversarial_competition/dataset.}} dataset, a commonly used subset of ImageNet. We selected two robust classifiers as most attacks achieved $0\%$ accuracy on standard classifiers. More specifically, we selected the RaWideResNet-101-2 by~\citet{peng2023robust} and WideResNet-50-2 by~\citet{salman2020adversarially}. We show the results for \oursacro{} and selected baselines, including Adversarial Content Attack~\citep{chen2023content} (ACA), in Tab.~\ref{tab:high_resolution}. The restricted baselines have a perturbation distance of $\ell_p=4/255$.
\begin{table*}[!ht]
\caption{Adversarial accuracy of \oursacro{} for robust classifiers on the ImageNet-Compatible dataset.}
\vspace{0.5em}
\label{tab:high_resolution}
    \scriptsize
    \centering
    \begin{tabular}{l c c} 
    \arrayrulecolor{black}
    & \citet{salman2020adversarially} & \citet{peng2023robust} \\
    \midrule
   \multicolumn{1}{l}{\cellcolor{Gray}\textbf{$\mathbf{\ell_\infty}$ restricted}} & \multicolumn{2}{c}{\cellcolor{Gray}}\\
    \hspace{3mm}FGSM~\citep{goodfellow2014explaining} & 58.8 & 66.5\\
    \hspace{3mm}DI-FGSM~\citep{xie2019improving} & 57.6 & 66.8\\
    \hspace{3mm}SI-NI-FGSM~\citep{lin2019nesterov} & 74.5 & 80.2\\
    \hspace{3mm}APGD~\citep{croce2020reliable} & 52.2 & 62.3\\
    \hspace{3mm}APGDT~\citep{croce2020reliable} & 46.5 & 59.1\\
    \multicolumn{1}{l}{\cellcolor{Gray}\textbf{$\mathbf{\ell_0}$ restricted}} & \multicolumn{2}{c}{\cellcolor{Gray}}\\
    \hspace{3mm}OnePixel~\citep{su2019one} & 85.3 & 88.2 \\
    \multicolumn{1}{l}{\cellcolor{Gray}\textbf{Unrestricted}} & \multicolumn{2}{c}{\cellcolor{Gray}}\\
     \hspace{3mm}CAA~\citep{hsiung2022caa} & 10.4 & 11.9 \\
   \hspace{3mm}PPGD~\citep{laidlaw2020perceptual} & 5.9 & 18.5 \\
    \hspace{3mm}LPA~\citep{laidlaw2020perceptual} & \textbf{1.6} & 8.8\\
    \hspace{3mm}DiffAttack~\citep{chen2023diffusion} & 6.0 & \underline{8.5} \\
    \hspace{3mm}ACA~\citep{chen2023content} & 4.6 & 6.5 \\
    \hspace{3mm}\oursacro{} (Ours) & \underline{2.5} & \textbf{4.1}\\
    \bottomrule
\end{tabular}
\end{table*}
As we can observe, \oursacro{} again achieves competitive performance, i.e., best and second-best accuracies. We show some examples of the unrestricted attacks in Fig.~\ref{fig:imagenet_compatible}. As expected, \oursacro{} preserves the semantics of the images and displays a high degree of realism. Surprisingly, the other diffusion-based attacks, DiffAttack and ACA, display more noticeable differences. ACA, in particular, has made major changes to the image.

\subsection{Purification of Common Corruptions}\label{sec:corruptions}
In addition to the purification of adversarial attacks, we test the applicability of \oursacro{} (GAP) on common corruptions~\citep{hendrycks2018benchmarking}. We show the robust accuracy of standard and robust classifiers before and after purification using DiffPure and GAP in Tab.~\ref{tab:corruptions}.
\begin{table*}[t!]
\caption{CIFAR10 robust accuracy of different adversarial training and purification methods for common corruptions on CIFAR10. If multiple threat models exist, we denote results as $\ell_\infty$/$\ell_2$. Best scores are bolds.}
\label{tab:corruptions}
\scriptsize
\begin{center}
  \begin{tabular}{l c c c c}
    Model & Base & DiffPure~\citep{nie2022DiffPure} &  \oursacro{}-GAP (Ours) & Architecture \\
    \toprule
    \hspace{3mm}Standard & 75.56$\pm$0.41 & 81.85$\pm$0.59 & \textbf{83.47$\pm$1.25} & WRN-28-10\\
\multicolumn{1}{l}{\cellcolor{Gray}\textbf{Adversarial Training}} & \multicolumn{4}{c}{\cellcolor{Gray}}\\
\hspace{3mm}\citep{cui2023decoupled} & 81.90 & \textbf{82.76} & 82.32 &  WRN-28-10\\
\hspace{3mm}\citep{wang2023better} & 81.38 / 87.96 & \textbf{81.98} / 86.40 & 81.58 / \textbf{88.74} & WRN-28-10\\
\hspace{3mm}\citep{wang2023better} & 83.90 / 89.24 & 84.16 / 87.14 & \textbf{84.30 / 89.86} & WRN-70-16 \\
\hspace{3mm}\citep{peng2023robust} & 83.32 & 83.76 & \textbf{83.94} & RaWRN-70-16 \\
      \bottomrule
  \end{tabular}
\end{center}
\vspace{-2.5em}
\end{table*} 
Our results show that \oursacro{} consistently increases the robust accuracy over the base model. In 5/7 settings, \oursacro{} achieves a better accuracy than DiffPure. Surprisingly, purifying the standard model makes it competitive with adversarially trained models, implying that purification does only benefit little when combined with a robust classifier.

\subsection{Large Perturbation Norms for restricted adversarial attacks}
\begin{figure*}[h]
    \centering
    \subfigure[APGD ($\ell_2$)]{\includegraphics[width=0.25\textwidth]{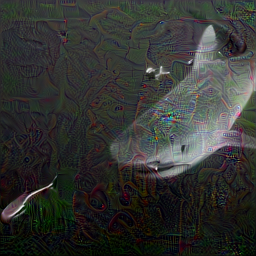}} 
    \subfigure[APGDT ($\ell_2$)]{\includegraphics[width=0.25\textwidth]{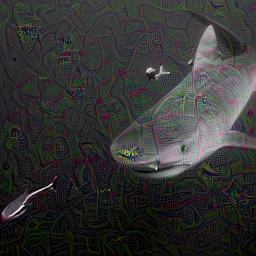}} 
    \subfigure[Square ($\ell_2$)]{\includegraphics[width=0.25\textwidth]{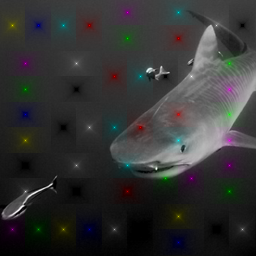}} 
    \subfigure[APGD ($\ell_\infty$)]{\includegraphics[width=0.25\textwidth]{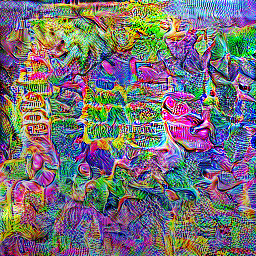}} 
    \subfigure[APGDT ($\ell_\infty$)]{\includegraphics[width=0.25\textwidth]{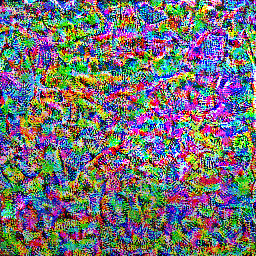}}
    \subfigure[Square ($\ell_\infty$)]{\includegraphics[width=0.25\textwidth]{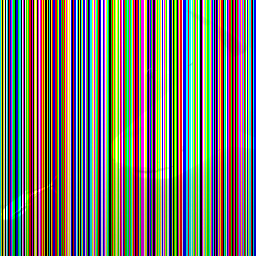}}
    \caption{Different adversarial attacks for the example in Fig.~\ref{fig:main}. The $\ell_\infty$ and $\ell_2$ distances are 188/255 and 18.47, respectively. All methods display major changes in the images compared to the original.}
    \label{fig:baseline_examples}
\end{figure*}

In Fig.~\ref{fig:baseline_examples}, we show adversarial examples of different attacks for the image in Fig.~\ref{fig:main}. We use the same distances \oursacro{} achieves.

\newpage
\subsection{Generative Adversarial Synthesis}\label{sec:GAS_images}
\begin{figure*}[h]
 \begin{minipage}{\linewidth}
     \centering
\includegraphics[width=\imagenetwidth px]{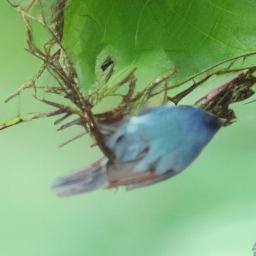}
\includegraphics[width=\imagenetwidth px]{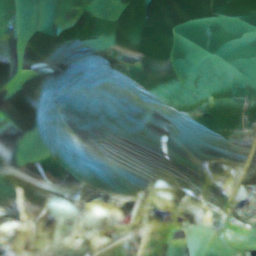}
\includegraphics[width=\imagenetwidth px]{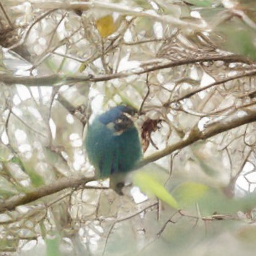}
\includegraphics[width=\imagenetwidth px]{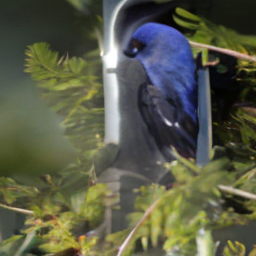}
\includegraphics[width=\imagenetwidth px]{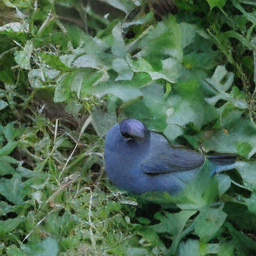}
\includegraphics[width=\imagenetwidth px]{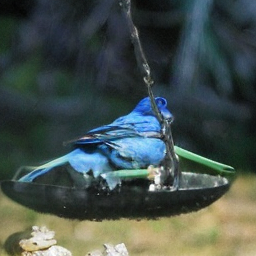}\\
\includegraphics[width=\imagenetwidth px]{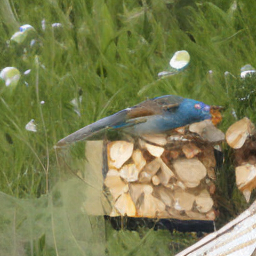}
\includegraphics[width=\imagenetwidth px]{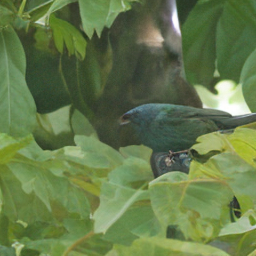}
\includegraphics[width=\imagenetwidth px]{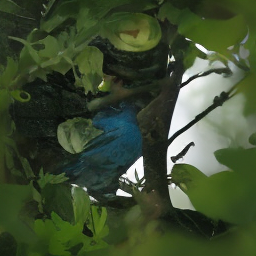}
\includegraphics[width=\imagenetwidth px]{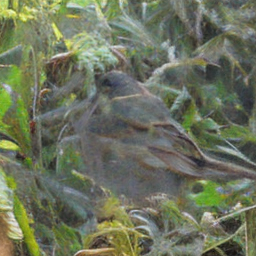}
\includegraphics[width=\imagenetwidth px]{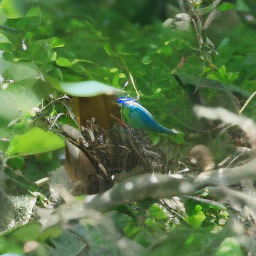}
\includegraphics[width=\imagenetwidth px]{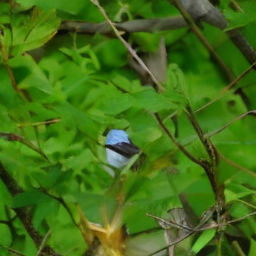}\\
\includegraphics[width=\imagenetwidth px]{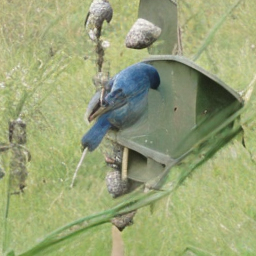}
\includegraphics[width=\imagenetwidth px]{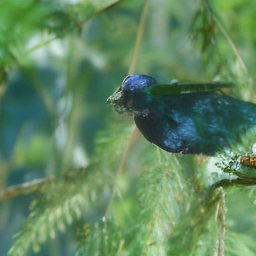}
\includegraphics[width=\imagenetwidth px]{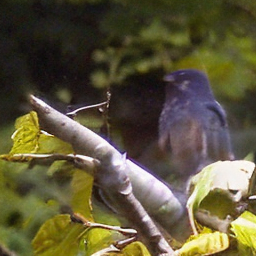}
\includegraphics[width=\imagenetwidth px]{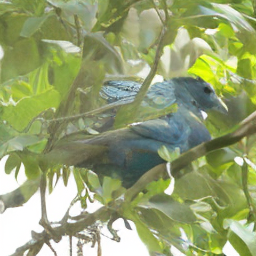}
\includegraphics[width=\imagenetwidth px]{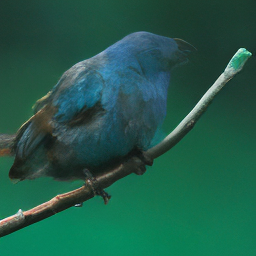}
\includegraphics[width=\imagenetwidth px]{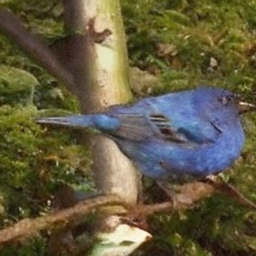}\\
\includegraphics[width=\imagenetwidth px]{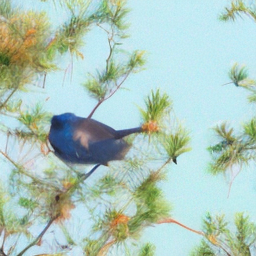}
\includegraphics[width=\imagenetwidth px]{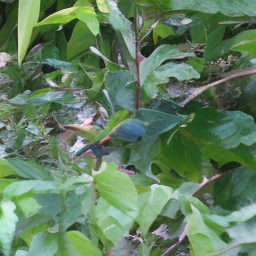}
\includegraphics[width=\imagenetwidth px]{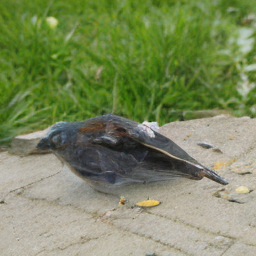}
\includegraphics[width=\imagenetwidth px]{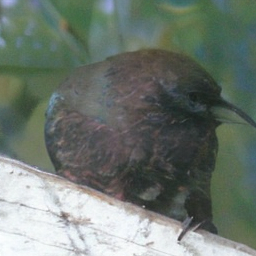}
\includegraphics[width=\imagenetwidth px]{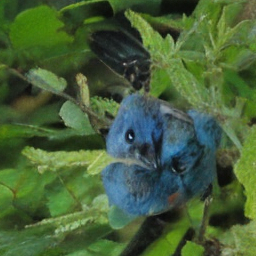}
\includegraphics[width=\imagenetwidth px]{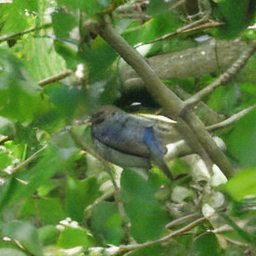}\\
\includegraphics[width=\imagenetwidth px]{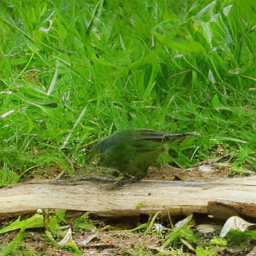}
\includegraphics[width=\imagenetwidth px]{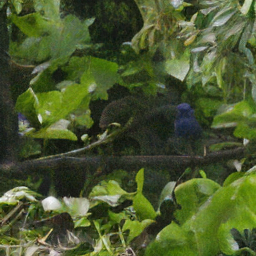}
\includegraphics[width=\imagenetwidth px]{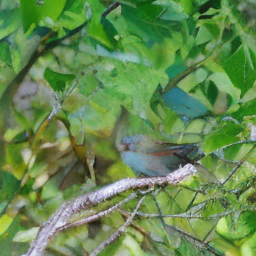}
\includegraphics[width=\imagenetwidth px]{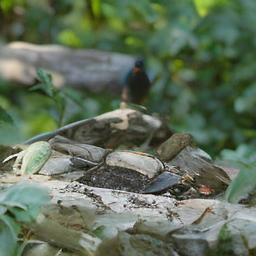}
\includegraphics[width=\imagenetwidth px]{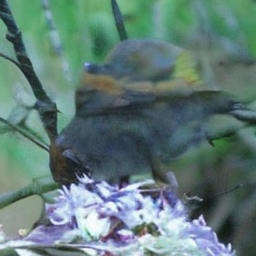}
\includegraphics[width=\imagenetwidth px]{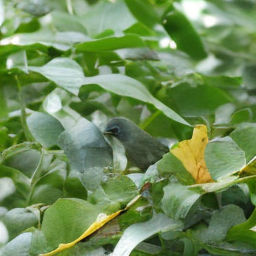}\\
\includegraphics[width=\imagenetwidth px]{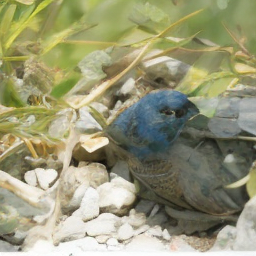}
\includegraphics[width=\imagenetwidth px]{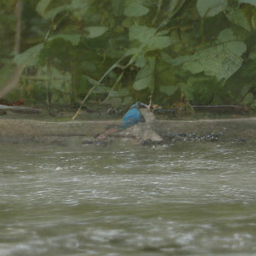}
\includegraphics[width=\imagenetwidth px]{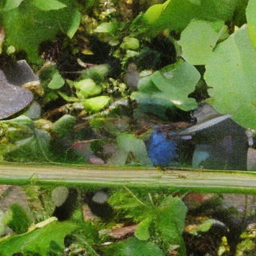}
\includegraphics[width=\imagenetwidth px]{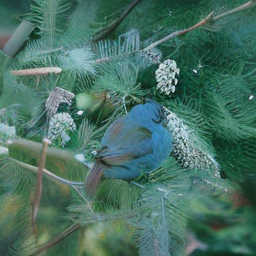}
\includegraphics[width=\imagenetwidth px]{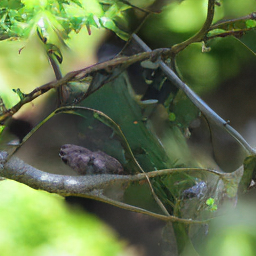}
\includegraphics[width=\imagenetwidth px]{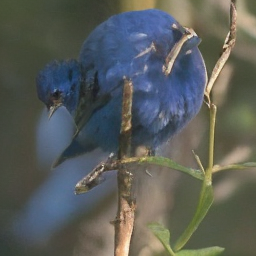}
\end{minipage}
\vspace{-0.5em}
\caption{Selected synthetic adversarial examples on ImageNet for the class "indigo bunting". All images display a high degree of realism and are classified wrongly into various 
\label{fig:imagenet_synthesis}
classes.}
\end{figure*}

In Fig.~\ref{fig:imagenet_synthesis}, we provide additional examples of the GAS task. The images are synthetic adversarial samples of the ImageNet class ``indigo bunting". While all images are classified wrongly, most of them contain the right core-semantics and display a high degree of realism.

\newpage
\subsection{Generative Adversarial Transformation}\label{sec:GAA_images}
\begin{figure}[ht]
 \begin{minipage}{\linewidth}
    \centering
\includegraphics[width=\imagenetwidth px]{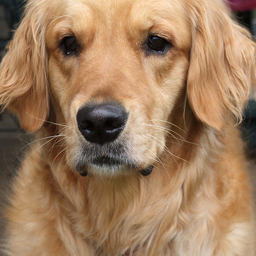}\hspace{0.75em}
\includegraphics[width=\imagenetwidth px]{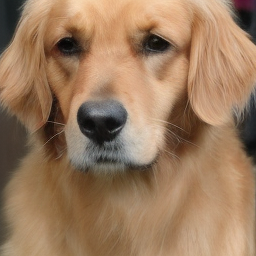}
\includegraphics[width=\imagenetwidth px]{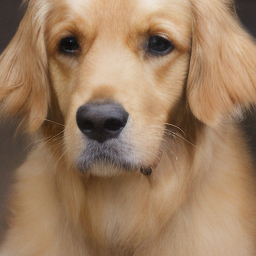}
\includegraphics[width=\imagenetwidth px]{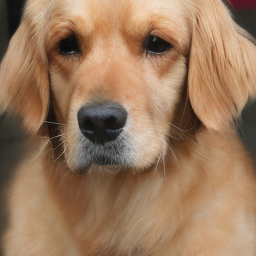}
\includegraphics[width=\imagenetwidth px]{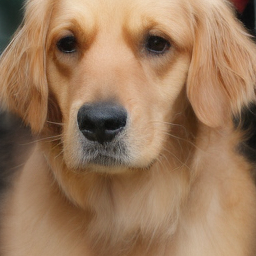}
\includegraphics[width=\imagenetwidth px]{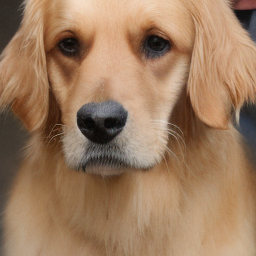}\\
\includegraphics[width=\imagenetwidth px]{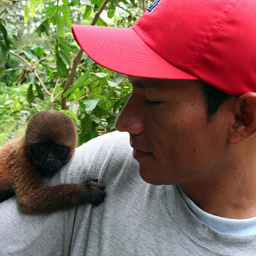}\hspace{0.75em}
\includegraphics[width=\imagenetwidth px]{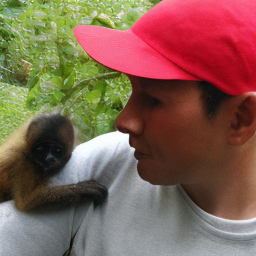}
\includegraphics[width=\imagenetwidth px]{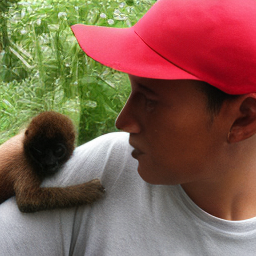}
\includegraphics[width=\imagenetwidth px]{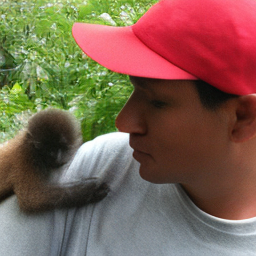}
\includegraphics[width=\imagenetwidth px]{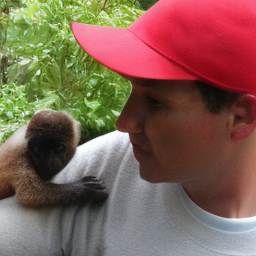}
\includegraphics[width=\imagenetwidth px]{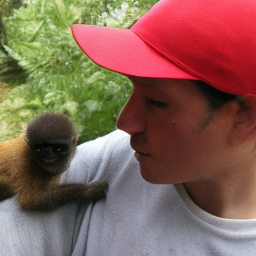}\\
\includegraphics[width=\imagenetwidth px]{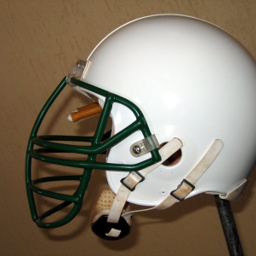}\hspace{0.75em}
\includegraphics[width=\imagenetwidth px]{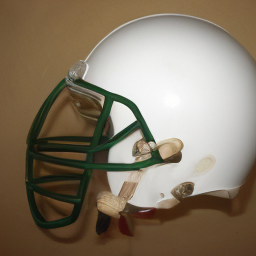}
\includegraphics[width=\imagenetwidth px]{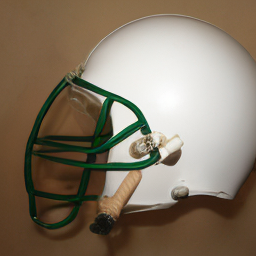}
\includegraphics[width=\imagenetwidth px]{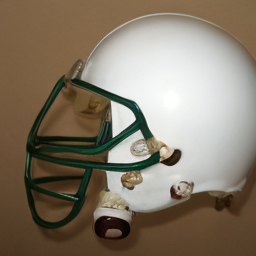}
\includegraphics[width=\imagenetwidth px]{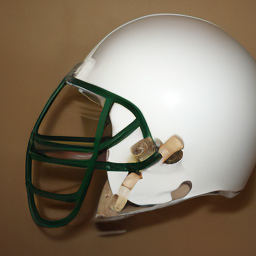}
\includegraphics[width=\imagenetwidth px]{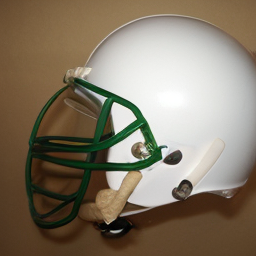}\\
\includegraphics[width=\imagenetwidth px]{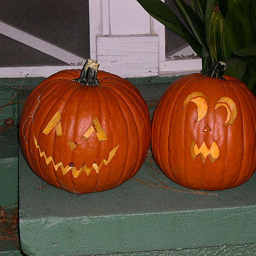}\hspace{0.75em}
\includegraphics[width=\imagenetwidth px]{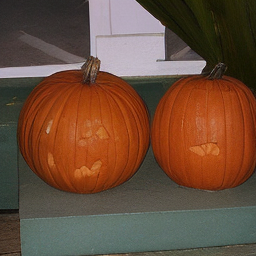}
\includegraphics[width=\imagenetwidth px]{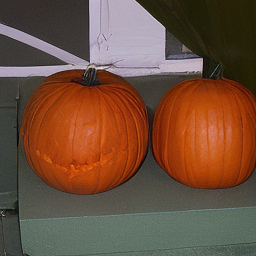}
\includegraphics[width=\imagenetwidth px]{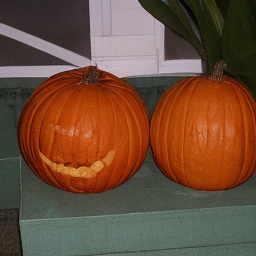}
\includegraphics[width=\imagenetwidth px]{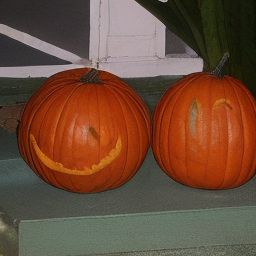}
\includegraphics[width=\imagenetwidth px]{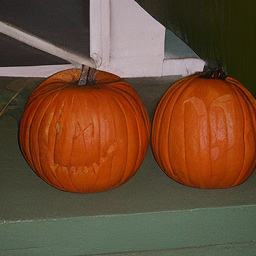}\\
\includegraphics[width=\imagenetwidth px]{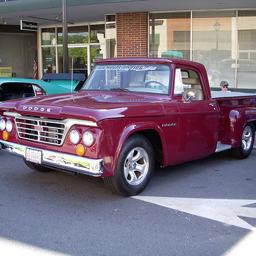}\hspace{0.75em}
\includegraphics[width=\imagenetwidth px]{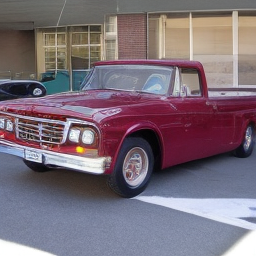}
\includegraphics[width=\imagenetwidth px]{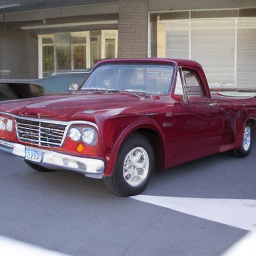}
\includegraphics[width=\imagenetwidth px]{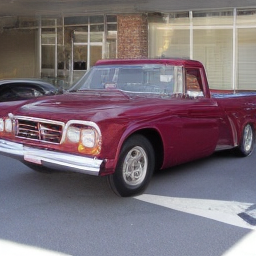}
\includegraphics[width=\imagenetwidth px]{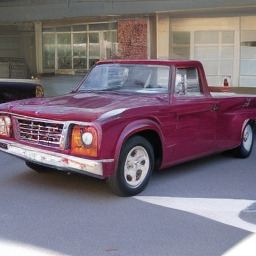}
\includegraphics[width=\imagenetwidth px]{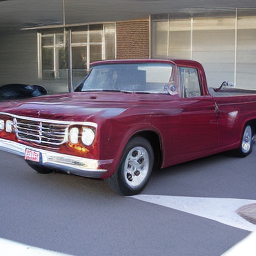}\\
\includegraphics[width=\imagenetwidth px]{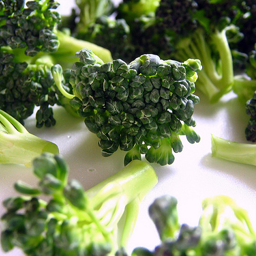}\hspace{0.75em}
\includegraphics[width=\imagenetwidth px]{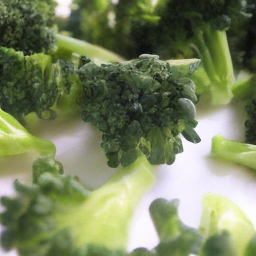}
\includegraphics[width=\imagenetwidth px]{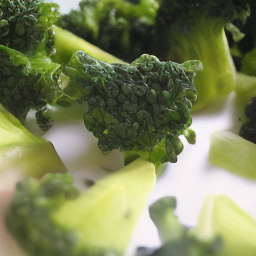}
\includegraphics[width=\imagenetwidth px]{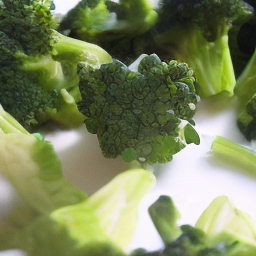}
\includegraphics[width=\imagenetwidth px]{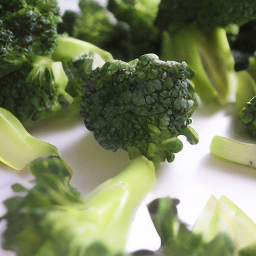}
\includegraphics[width=\imagenetwidth px]{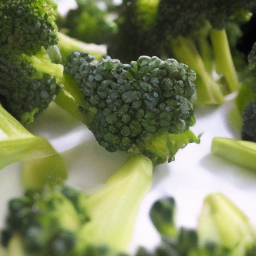}
\end{minipage}
    \centering
\subfigure[\label{fig:transform_left}Original.]{\hspace{.14\textwidth}} 
    \subfigure[\label{fig:transform_right}Adversarial Examples.]{\hspace{.70\textwidth}}
\vspace{-0.5em}
\caption{Selected transformed adversarial examples on ImageNet. While the adversarial examples are classified wrongly, the original images are classified correctly. All images maintain the semantics while being outside of common perturbation norms.}
\label{fig:imagenet_transform}
\end{figure}

In Fig.~\ref{fig:imagenet_transform}, we show additional examples of the GAT task. All original images are classified correctly into the ImageNet classes ``golden retriever", ``spider monkey", ``football helmet", ``jack-o'-lantern", ``pickup truck", and ``broccoli", while the adversarial images are classified as ``cocker spaniel", ``gibbon", ``crash helmet", ``barrel", ``convertible", and ``custard apple", respectively. While all adversarial images display subtle differences they do not alter the core semantics of the images and are not captured by common $\ell_p$-norms. 

\subsection{Runtime comparison of the attacks.}\label{sec:runtimes}
All experiments were conducted on A100s. In Tab.~\ref{tab:runtimes}, we report the runtimes in seconds of various methods. The numbers display the average time to generate one adversarial example on the ImageNet-Compatible dataset. \change{Note that sampling an image without guidance using the same generative model as \oursacro{} takes $15.00$ seconds. The difference stems from the additional overhead induced by the gradient computations.}
\begin{table}[h!]
    \scriptsize
    \centering
    \caption{Average runtimes in seconds of the different attacks on an A100 to generate one adversarial images for the ImageNet-Compatible dataset.}
    \resizebox{\textwidth}{!}{
    \begin{tabular}{ccccccccccccc}
         FGSM & DIFGSM & SINIFGSM & Square & FAB & APGD & APGDT & OnePixel & LPA & PPGD & DiffAttack & ACA &\oursacro{} \\
         \midrule
0.45 & 0.18 & 0.68 & 36.04 & 74.39 & 0.40 & 1.86 & 0.64 & 0.96 & 0.50 & 19.14 & 188.80 & 79.34\\
         \bottomrule
    \end{tabular}}
    \label{tab:runtimes}
\end{table}

\end{document}